\documentclass{article}

\usepackage{microtype}
\usepackage{graphicx}
\usepackage{subfigure}
\usepackage{booktabs} 

\usepackage{hyperref}

\usepackage[accepted]{icml2025}

\usepackage{amsmath}
\usepackage{amssymb}
\usepackage{mathtools}
\usepackage{amsthm}

\usepackage{hyperref}
\usepackage{url}
\usepackage{graphicx} 
\usepackage{amssymb}
\usepackage{booktabs}
\usepackage{pifont}
\usepackage{xcolor}
\usepackage{multirow}

\usepackage[capitalize,noabbrev]{cleveref}

\theoremstyle{plain}

\theoremstyle{definition}

\theoremstyle{remark}

\usepackage[textsize=tiny]{todonotes}

\icmltitlerunning{Unreal-MAP: Unreal-Engine-Based General Platform for Multi-Agent Reinforcement Learning}

\begin{document}

\twocolumn[

\icmltitle{Unreal-MAP: Unreal-Engine-Based General Platform  \\
           for Multi-Agent Reinforcement Learning}



\icmlsetsymbol{equal}{*}

\begin{icmlauthorlist}
\icmlauthor{Tianyi Hu}{ucas,casia}
\icmlauthor{Qingxu Fu}{ucas,ali}
\icmlauthor{Zhiqiang Pu}{ucas,casia}
\icmlauthor{Yuan Wang}{ucas,casia}
\icmlauthor{Tenghai Qiu}{casia}

\end{icmlauthorlist}

\icmlaffiliation{ucas}{School of Artificial Intelligence, University of Chinese Academy of Sciences, Beijing}
\icmlaffiliation{casia}{Institute of Automation, Chinese Academy of Sciences, Beijing}
\icmlaffiliation{ali}{Alibaba (China) Co., Ltd., Beijing}

\icmlcorrespondingauthor{Zhiqiang Pu}{zhiqiang.pu@ia.ac.cn}

\icmlkeywords{Machine Learning, ICML}

\vskip 0.3in
]



\printAffiliationsAndNotice{}  

\begin{abstract}

In this paper, we propose \textit{Unreal Multi-Agent Playground} (Unreal-MAP), an MARL general platform based on the Unreal-Engine (UE). Unreal-MAP allows users to freely create multi-agent tasks using the vast visual and physical resources available in the UE community, and deploy state-of-the-art (SOTA) MARL algorithms within them. Unreal-MAP is user-friendly in terms of deployment, modification, and visualization, and all its components are open-source. We also develop an experimental framework compatible with algorithms ranging from rule-based to learning-based provided by third-party frameworks.  Lastly, we deploy several SOTA algorithms in example tasks developed via Unreal-MAP, and conduct corresponding experimental analyses. We believe Unreal-MAP can play an important role in the MARL field by closely integrating existing algorithms with user-customized tasks, thus advancing the field of MARL. The source code for our project is available at \url{github.com/binary-husky/unreal-map} and \url{github.com/binary-husky/hmp2g}.

\end{abstract}

\section{Introduction}

\begin{figure}[!h] 
    \centering
    \includegraphics[width=1.0\linewidth]
    {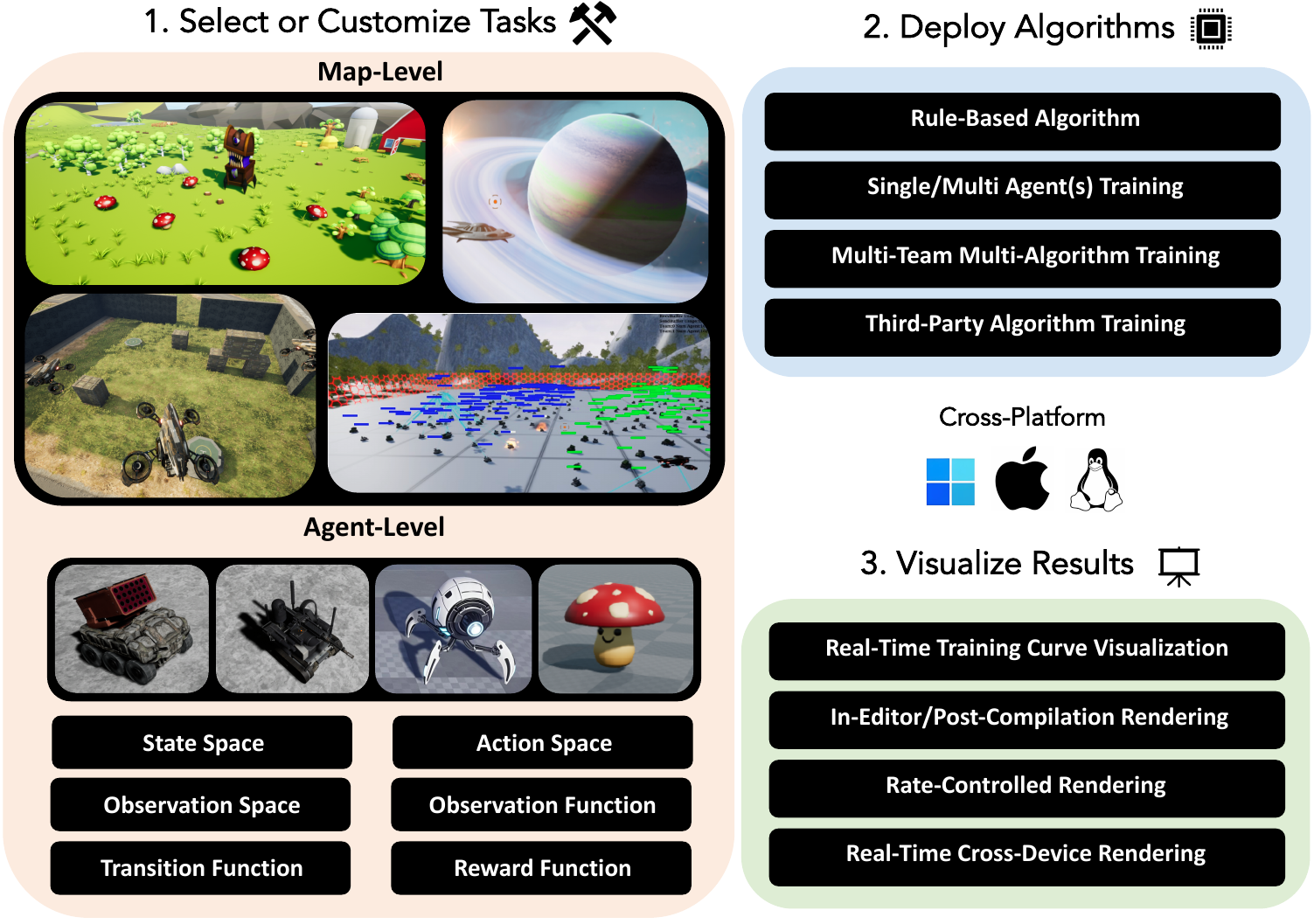}
    \caption{The research workflow for using Unreal-MAP. For novice users, Unreal-MAP provides direct access to built-in tasks, and offers comprehensive algorithm deployment functions and result visualization capabilities. For advanced users, Unreal-MAP enables the modification of built-in tasks or the creation of new tasks to test research ideas.}
    \label{fig:Use_Unreal-MAP}
\end{figure}

Multi-agent reinforcement learning (MARL) has demonstrated remarkable potential in many practical fields, including swarm robotic control~\citep{kalashnikov2018scalable, chen2020autonomous}, 
autonomous vehicles~\citep{autonomous_vehicles_1}, and video games~\citep{vinyals2019grandmaster}. There are many classical and practical algorithms that have emerged in the field of MARL, such as QMIX~\cite{QMIX}, QPLEX~\cite{QPLEX}, MAPPO~\cite{MAPPO} and HAPPO~\cite{HARL}. The development and success of these algorithms could not be achieved without the support of numerous well-designed simulation environments. These environments provide ample simulated data for testing and comparing algorithms, thereby fostering a variety of more advanced algorithms.

Obviously, the progression of MARL algorithms will not stop at mere comparison of performance; their ultimate aspiration lies in the deployment within practical applications. However, when considering rapid deployment in real-world applications, current popular simulation environments cannot be freely and effectively customized according to user needs. Additionally, crafting new domain-specific environments for particular problems frequently encounters challenges in swiftly integrating with existing MARL algorithm libraries\footnote{We refer to MARL libraries as experimental frameworks, characterized by their interfaces for both common environments and various algorithms.}, thus hindering the deployment of state-of-the-art (SOTA) algorithms to meet practical requirements~\cite{2023review}.

The existence of modern game engines offers a possible solution to the aforementioned issues. Game engines are software frameworks originally designed to simplify the development of video games~\cite{boyd2017reinforcement}. Over years of evolution, modern game engines have established vast development communities, along with rich rendering resources and physics engine materials~\cite{wheeler2023reinforcement}. If possible, these potential resources could greatly facilitate the application deployment within the MARL domain. Moreover, game engines are closely linked with the rapidly developing generative AI, which has the potential to quickly transform user needs into real products. Some works have utilized simulation data from game engines for the training of generative AI~\cite{lu2024genex}, whereas others have established a feedback loop, incorporating these trained generative models into game engine plugins to facilitate scene creation~\cite{nvidia_ace}. This suggests that combining game engines with MARL could have a very promising future.

However, there remains a significant gap between the game development community associated with game engines and the MARL community. Although some efforts, such as URLT~\cite{URLT} and Unity-ML Agents~\cite{Unity}, have built frameworks for freely constructing RL training scenarios using game engines\footnote{In this paper, we refer to the works that allow for the free construction of RL training scenarios as general platforms~\cite{Unity}.}, they still face issues in training efficiency and scaling complexity~\cite{kaup2024review}. To address this, we propose Unreal-MAP, an MARL general platform based on the Unreal Engine\footnote{\url{https://www.unrealengine.com}.}. Compared to existing solutions, Unreal-MAP possesses the following main features:

\textbf{(1) Fully Open-Source and Easily Modifiable}, Unreal-MAP utilizes a layered design, and all components from the bottom-level engine to the top-level interfaces are open-sourced. Users can easily modify all elements of an MARL task by focusing only on the higher-level operational layers.
\textbf{(2) Optimized Specifically for MARL}, the underlying engine of Unreal-MAP has been optimized to enhance efficiency in large-scale agent simulations and data transmission. This optimization allows users to develop simulations with heterogeneous, large-scale, multi-team settings that showcase distinctive multi-agent features through Unreal-MAP.
\textbf{(3) Parallel Multi-Process Execution and Controllable Single-Process Time Flow}, Unreal-MAP supports the parallel execution of multiple simulation processes as well as the adjustment of the simulation time flow speed in a single process. Users can accelerate simulations to speed up training or decelerate simulations for detailed slow-motion analysis.

To fully utilize the capabilities of Unreal-MAP, we also develop an MARL experimental framework known as the Hybrid Multi-Agent Playground (HMAP). This framework includes implementations of 
rule-based algorithms, built-in learning-based algorithms, and algorithms from third-party frameworks such as PyMARL2~\citep{pymarl2} and HARL~\citep{HARL}. 
By leveraging Unreal-MAP and HMAP, users can rapidly customize environments and deploy algorithms, validate new research ideas, and apply them in practical scenarios. The overview of the research workflow for using Unreal-MAP is depicted in Figure~\ref{fig:Use_Unreal-MAP} and more details can be found in Appendix~\ref{WorkFlow}.

The contributions of this work are summarized as follows: firstly, an MARL general platform based on the Unreal Engine; secondly, an accompanying modular MARL experimental framework; thirdly, a collection of highly extensible example scenarios based on Unreal-MAP and related experimental analyses. We believe Unreal-MAP can serve as a comprehensive tool to advance the development of MARL and ultimately facilitate their application in real-world scenarios.

\section{Related Work}

\textbf{Simulation Environments for MARL.} Existing MARL environments can be broadly divided into \textit{domain-specific environments} and \textit{environment-suites}. The former includes a series of tasks designed around the same domain, which usually share common genre and similar benchmark metrics. Notable examples of these works include MPE~\cite{MPE}, SMAC~\cite{SMAC}, GRF~\cite{GRF}, MAMuJoCo~\cite{MAMuJoCo}, and GoBigger~\cite{GoBigger}. Due to architectural constraints, domain-specific environments cannot be freely modified according to user needs. For instance, a StarCraft-based SMAC scenario can never be reconfigured into soccer-style tasks. On the other hand, \textit{environment suites} consist of sets of environments packaged together, commonly used to more conveniently benchmark the performance of algorithms. Some works also redesign built-in environments to enhance training speed. Typical works in this category include PettingZoo~\cite{Pettingzoo} and JaxMARL~\cite{JaxMARL}. Nevertheless, these suites are just collections of domain-specific environments and thus lack flexible modification capabilities. More details of related MARL environments can be found in Appendix~\ref{Sec:related work and comparison}.

\textbf{General Platforms for RL.} These kinds of works enable users to create environments with arbitrary visual or physical complexity and deploy existing RL algorithms. Unity-ML Agents \cite{Unity} is a mature general platform built on the Unity engine. It allows users to develop new scenes using the Unity editor and train them using Python-based RL algorithm libraries. However, its limited open-source implementation hinders further customization~\cite{wheeler2023reinforcement}, and there are still issues in scaling complexity and simulation fidelity~\cite{kaup2024review}. Compared to Unity, Unreal Engine (UE) offers full open-source access and lower learning curves\footnote{The UE editor employs the graphical programming language Blueprints, offering a lower learning curve than its native C++ library, even for those familiar with C++~\cite{boyd2017implementing}.}~\cite{boyd2017implementing}, and is more conducive to the development of a general platform. URLT~\cite{URLT} is an RL general platform based on UE, but its algorithm part relies on Blueprint scripts and is difficult to integrate with existing cutting-edge algorithm libraries. Moreover, it does not support controllable time flow, resulting in very slow training and making it inapplicable to large-scale agent simulations. Currently, there is a lack of a powerful, MARL-targeted general platform.

\textbf{Generative Physics Engines for RL.} Emerging generative physics engines show potential for translating user needs into actual products, with the ultimate goal of generating reliable simulations based on user prompts. Research in this area is still in its infancy. A typical work in this domain is Genesis~\cite{Genesis}, which can generate physical scenes for RL training through prompts and supports differentiable
scenes. However, it still relies on underlying physical engines and 3D models, which is not fully generative. Emerging AI-driven generative physics engines may replace current lightweight physical engines~\cite{kaup2024review}, but they still depend on the integration with game engines to achieve enhanced generative workflow. Specifically, game engines can provide generative AI with enough data for training~\cite{lu2024genex}, while generative AI can assist game engines in quickly developing scene-related assets, such as 3D models or physical engines. In fact, NVIDIA has proposed a work for high-quality 3D asset generation~\cite{Edify_3D}, and provided a generative AI plugin for use with UE~\cite{nvidia_ace}. Combining game engine with MARL naturally aligns with the development trend of generative AI.

\section{Background}
\label{sec:Background}

To accommodate various interaction relationships among multi-agent and multi-team scenarios~\citep{Fuzzy}, we use Partially Observable Markov Game (POMG)~\citep{Markov_game, POMG_survey} to model the MARL problem. A POMG can be represented by an 8-tuple $\langle N, 
\{S^i\}_{i\in N}, \{O^i\}_{i\in N}, \{\Omega^i\}_{i\in N},
\{A^i\}_{i\in N}, \{\mathcal{T}^i\}_{i\in N}, r, \gamma \rangle$. $N$ is the set of all agents, $\{S^i\}_{i\in N}$ is the global state space which can be factored as $\{S^i\}_{i\in N} =\times_{i\in N} S^{(i)} \times S^{E}$, where $S^{(i)}$ is the state space of an agent $i$, and $S^{E}$ is the environmental state space, corresponding to all the non-agent entities. $\{O^i\}_{i\in N}=\times_{i\in N} O^{(i)}$ is the joint observation space and $\{\Omega^i\}_{i\in N}$ is the set of observation functions. Similarly, $\{A^i\}_{i\in N}$ is the joint action space of all agents. $\{\mathcal{T}^i\}_{i\in N}$ is the collection of all agents' transitions and the environmental transition. Finally, $\gamma$ is the discount factor and $r: \{S^i\}_{i\in N} \times \{A^i\}_{i\in N} \times N \rightarrow \mathbb{R}$ is the agent-level reward function.

We define \textit{team} as a collection of agents, which all share the same overall goal in a purely cooperative form. 
Agents within the same team aim to find an optimal joint policy that maximizes the cumulative reward for the whole team. Denoting the joint policy of a certain team $A \subseteq N$ as $\bar{\pi}_A$, the optimal policy $\bar{\pi}^{*}_A$ can be represented as:
\begin{equation}
\label{Eqa: Joint policy}
\bar{\pi}^{*}_A = \arg\max_{\bar{\pi}_A} \mathbb{E}_{\bar{\pi}_A} 
\left[
    \sum_{k=0}^{\infty} \gamma^k \sum_{i \in A} r^i_{t+k} \mid \bar{s}_t = \bar{s} 
\right],
\end{equation}
where $\bar{s}$ is the initial global state, $\gamma^k \sum_{i \in A} r^i_{t+k}$ is the discounted return of team $A$, $r^i_{t+k}$ is the reward of an agent $i \in A$ at timestep $t+k$.

\section{Unreal-MAP}

\begin{figure*}[!h] 
    \centering
    \includegraphics[width=0.9\linewidth]
    {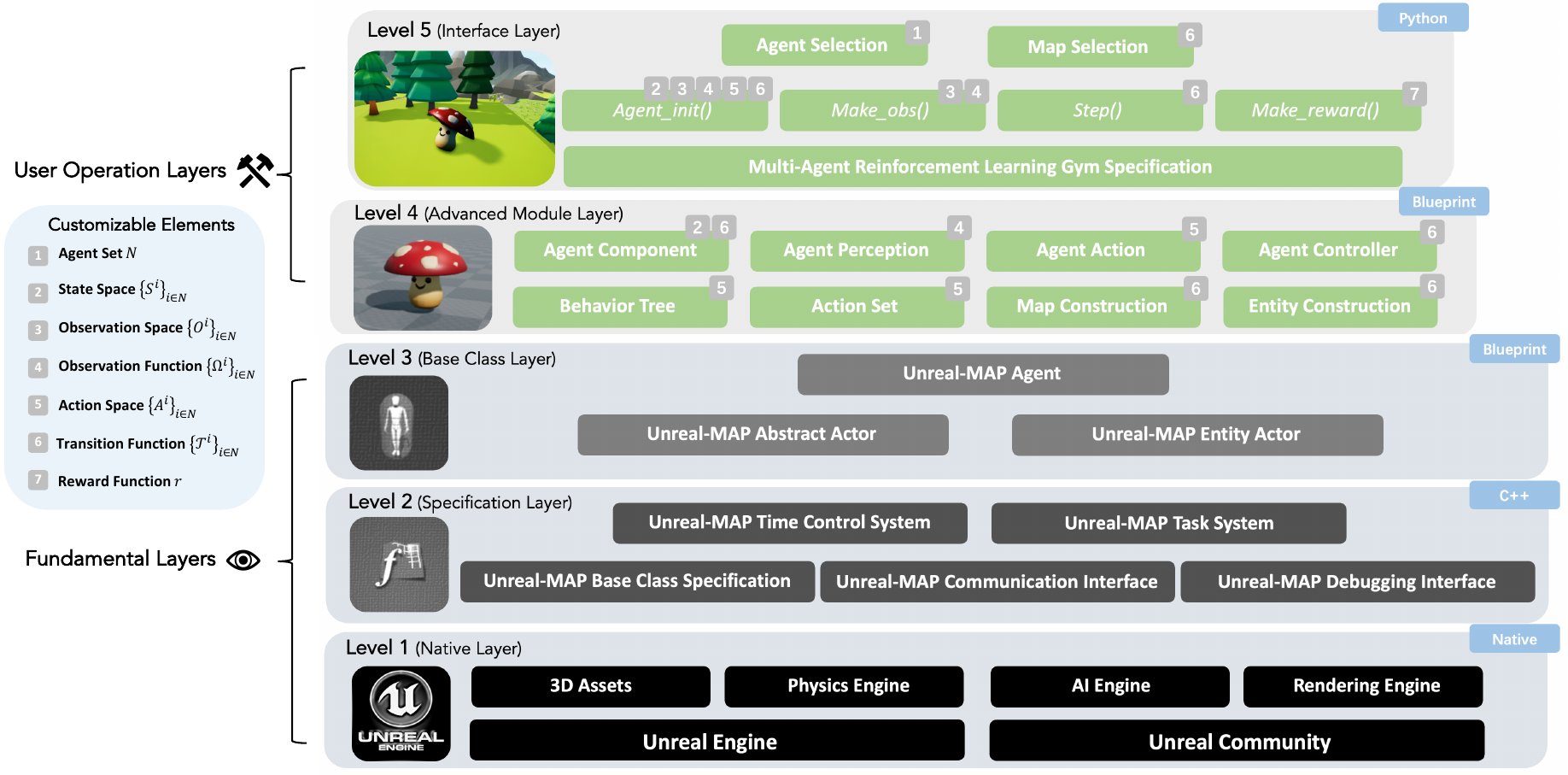}
    \caption{Architecture of Unreal-MAP. Unreal-MAP employs a hierarchical, five-layered architecture, all of which are open source. Users can modify all elements within POMG by configuring parameters through the Python-based \textit{interface layer}. For more advanced development requirements, users can conveniently adjust scenario elements using Blueprint through the \textit{advanced module layer}.}
    \label{fig:Framework of Unreal-MAP}
\end{figure*}

\subsection{Basic Concepts in Unreal-MAP}


Multi-agent simulation can demonstrate great diversity in different domains. We introduce several new concepts that align with human intuition as well as the requirements of multi-agent simulation.

\textbf{Agents} and \textbf{Teams:} 
    Agents are the basic decision-making units in the environments.
    Unreal-MAP introduces a new concept \textit{team} to distinguish agents with different goals. Unreal-MAP supports numbers of teams, where teams may engage in competition or cooperation. Each team possesses its own independent goal and is equipped with a separate learning-based (or rule-based) algorithm.
    
\textbf{Entities:}
Entities are objects in simulation that do not make decisions but still has important functionality. For instance \textit{a street lamp} or \textit{a dynamic obstacle}. A shared characteristic of these objects is that they must be removed or reinitialized when an episode ends or a new episode starts.
    
\textbf{Tasks} and \textbf{Scenarios:}
Tasks corresponds to POMGs defined in Section~\ref{sec:Background}. The properties of tasks in Unreal-MAP include the types and numbers of agents, their team affiliations, as well as each agent's state space, action space, etc. A scenario can give rise to a series of tasks, which typically share similar reward functions, implying that the objectives to be achieved by the multi-agent systems are 
the same.

\textbf{Maps:}
Maps in Unreal-MAP determine where the task takes place. A map can be \textit{a small room}, or \textit{a city full of buildings.} It is a great advantage that Unreal-MAP decouples the concept of tasks and maps, as users can conveniently deploy a task in new maps (as long as the agent has the appropriate size and a suitable position initialization function).

\textbf{Events:}
We define an event system to simplify the reward crafting procedure. 
For instance, an event will be generated when an agent is destroyed or an episode is ended. When it is time to compute next-step reward, these events will provide convenient reference.

\subsection{Utilizing Unreal-MAP to customize tasks}

Unreal-MAP employs a hierarchical five-layer architecture, where each layer builds upon the previous one. From bottom to top, the five layers are: \textit{native layer}, \textit{specification layer}, \textit{base class layer}, \textit{advanced module layer}, and \textit{interface layer}. \textbf{Users only need to focus on the \textit{advanced module layer} and the \textit{interface layer}.} In most cases, modifying the \textit{interface layer} is sufficient to alter all elements of tasks.

Figure~\ref{fig:Framework of Unreal-MAP} shows the internal architecture of Unreal-MAP. Specifically, the \textit{native layer} includes assets from the Unreal community and the Unreal Engine, some part of which have been optimized for MARL compatibility. The \textit{specification layer} consists of Unreal-MAP's underlying systems and programming specifications, all implemented in C++. The \textit{base class layer} includes all basic classes implemented using Blueprints. These three layers, also known as the fundamental layers, form the foundation of Unreal-MAP, .

The top two layers of Unreal-MAP are user operation layers. The \textit{advanced module layer}, based on Blueprints, allows for the modification of agents' physical properties such as appearances, perceptions and kinematics, thereby enabling the development of various agents. This layer also facilitates the development of environmental entities and maps. The top layer is the \textit{interface layer}, implemented in Python and compliant with the gym standard. It supports customizable observations and reward functions, and allows for the selection of maps and agents.
More details about the architecture of Unreal-MAP  can be found in Appendix~\ref{App:Unreal-MAP Details}.


Thanks to the hierarchical architecture of Unreal-MAP, users can easily customize tasks through simple operations via top layers. Here we provide a detailed explanation of how each element of a task\footnote{Excluding the discount factor, which can be easily specified on the algorithm side.} is customized within Unreal-MAP.

\textbf{Agent Set.}  Within the interface level of Unreal-MAP, the agent selection module enables users to specify the types, numbers, and associated teams of agents.

\textbf{State Space.}
The global state is composed of the states of individual agents and the environmental state. Customization of the environmental state can be achieved by selecting different maps and modifying them along with related entities. The state of the agents can be customized through the \textit{agent\_init} function in the \textit{advanced module layer} and the agent component module in the \textit{interface layer}.

\textbf{Observation Space and Observation Function.}
Unreal-MAP transmits global information from the UE side to the Python side, where the \textit{make\_obs} function in the \textit{interface layer} is used to construct the agents' observations. Modification of this function allows for the customization of each agent's observation space and function. Moreover, modifying agents' properties, such as the observation range, can also change their observations. Additionally, Unreal-MAP supports more sophisticated agent observation simulation mechanisms, such as masking entities blocked by walls, which can be implemented through the agent perception module in the \textit{advanced module layer}.

\textbf{Action Space.}
Unreal-MAP supports continuous actions, discrete actions, and hybrid actions. Users can assign a built-in action set to each agent via the \textit{agent\_init} function in the \textit{interface layer}. Furthermore, a deeper customization of agent actions can be achieved through the agent action-related modules in the \textit{advanced module layer}.

\textbf{Transition Function.} 
Similar to the state space, the transition function in Unreal-MAP is comprised of local transitions of all agents and environmental transitions. The latter can be modified through map-related and entity-related modules. Local transitions of agents can be customized by modifying the \textit{agent\_init} function and the \textit{step} function, or more deeply through the agent component modules and agent controller modules, such as agent kinematics.

\textbf{Reward Function.}
Unreal-MAP constructs rewards using global information and an event system. Users can customize the agents' rewards by modifying the \textit{make\_reward} function, which supports team and individual rewards, as well as sparse and dense reward structures.

\subsection{Other Features of Unreal-MAP}

Unreal-MAP connects the UE community and MARL community. It allows users to utilize the extensive, realistic resources such as models, rendering materials and physical simulations from the UE community to develop scenarios for MARL training. This is the biggest advantage of Unreal-MAP over other MARL environments in terms of the highly scalability and realism of the created scenarios. Furthermore, this section will introduce other features of Unreal-MAP that are beneficial in the context of MARL.

\textbf{Computational efficiency.} Numerous modifications have been made to the underlying engine to adapt it for efficient MARL training. These include optimizations within the simulation engine and enhancements in the communication between the simulator and the algorithm side (details are provided in Appendix \ref{Underlying Optimization}). In practical training, Unreal-MAP also supports a non-render training mode without rendering frame computation.

\textbf{Controllable simulation time flow.} Unreal-MAP optimizes the time flow control mechanism in Unreal Engine (details are provided in Appendix \ref{Time}). Users can easily modify the time dilation factor to adjust the ratio of simulated time flow and real time flow. The ability to control the time flow offers numerous benefits. On one hand, users can accelerate the simulation time for rapid training or decelerate for debugging. On the other hand, since adjusting the speed of simulation does not influence memory resources, users can make fuller use of computational resources by adjusting the time dilation factor, with more detailed information available in Section~\ref{sec:Eff}.

\textbf{Compatibility with multiple systems and computational backends.} Unreal-MAP is natively compatible with creating environments and deploying algorithms on Windows, Linux, and MacOS. It supports training on pure CPU setups as well as on hybrid CPU/GPU configurations. Furthermore, Unreal-MAP supports cross-device real-time rendering\footnote{Remotely connecting to a non-render client running inside a server via network, and rendering the ongoing training process locally via TCP\&UDP.}, allowing users to conduct multi-process training on a Linux server while performing real-time rendering of specific processes on a Windows host.

\section{HMAP}

\begin{figure*}[!t]
    \centering
    \includegraphics[width=0.8\textwidth]
    {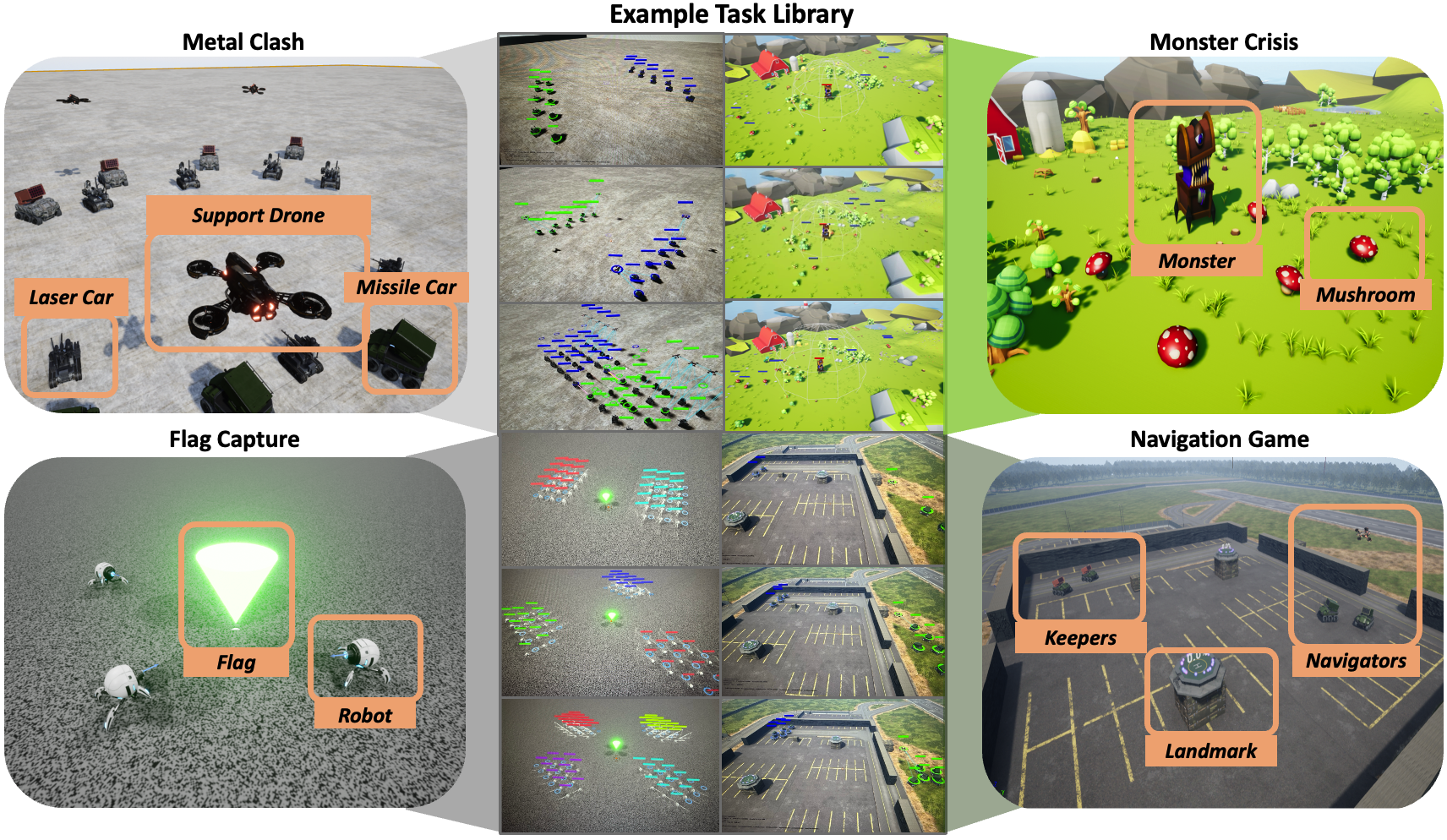}
    \caption{Example scenarios and tasks of Unreal-MAP. Users can develop new scenarios using Unreal-MAP, and create a variety of MARL tasks by adjusting properties such as the number, types, and teams of agents.}
    \label{fig:task_library}
\end{figure*}

To facilitate the deployment of algorithms for Unreal-MAP, we also develop an experimental framework HMAP. HMAP is a multi-agent experimental framework with decoupled \textit{Task-Core-Algorithm} components. 
Currently, HMAP not only integrates Unreal-MAP's tasks, but also supports other MARL environments such as SMAC~\citep{SMAC} and MPE~\citep{MPE}. On the algorithm side, HMAP also supports a wide range of algorithms. This includes rule-based algorithms (most of them are built-in policies for Unreal-MAP example tasks), single-agent RL algorithms like DQN~\citep{DQN} and SAC~\citep{SAC}, as well as MARL algorithms such as MAPPO~\citep{MAPPO} and HAPPO~\citep{HARL}. Furthermore, HMAP is compatible with third-party frameworks, supporting all algorithms from PyMARL2~\citep{pymarl2} and HARL~\citep{HARL}.

\textbf{The unique feature of HMAP is its support for multi-team training.} By thoroughly decoupling algorithms from tasks, HMAP employs its core as a \textit{``glue module"}, enabling any algorithm module to control teams within any task module. The highly modular design presents three key benefits. 
Firstly, it enables modification of built-in policies in tasks within Python-based algorithm modules, which can significantly reduce the workload of building non-learning-based policy\footnote{For example, the policy of a Non-Player Character (NPC).} on the UE side. Secondly, it enables teams controlled by multiple algorithms to interact within a same task, facilitating co-training of algorithms from different frameworks under the same task. Thirdly, it is user-friendly, as all experimental configurations based on HMAP can be implemented through a single JSON file. After completing the configuration, users can initiate the training task with just one line of code. More details of HMAP can be found in Appendix~\ref{app:HMAP Details}.

\section{Example Scenarios and Tasks}
\label{sec:Scenarios and Tasks}

Unreal-MAP includes a variety of basic scenarios for multi-agent systems, each of which is extensible and can be used to create numerous tasks. This section describes 4 example scenarios, which are used to develop 15 tasks applied in Section~\ref{sec:Experiments}. We use these example scenarios to demonstrate that Unreal-MAP can be used to construct tasks with distinct multi-agent characteristics. These characteristics include heterogeneity, large-scale, multi-team, sparse team rewards, and multi-agent games. More details of these example scenarios can be referred to Appendix~\ref{app:Scenario Details}.

\textbf{\textit{Metal Clash}} - Scenario designed for heterogeneous and large-scale multi-agent tasks. It involves an SMAC-style competition between two teams of agents. Each team can be controlled by either rule-based or learning-based algorithms.
\textit{Metal Clash} provides three types of basic agents: missile cars, laser cars and support drones. The properties of each basic agent, such as maximum speed and health points (HP), are encapsulated as configurable parameters. Users can easily modify these parameters, creating a variety of heterogeneous agent types beyond the original three. The number and types of agents in each team can be freely changed, altering the features and difficulty of the tasks.

\textbf{\textit{Monster Crisis}} - Scenario designed for sparse team rewards in a multi-agent cooperative setting. This is a village-style scenario where several mushroom agents need to resist the invasion of a monster. The entire team receives a positive reward only if the they kill the monster, and there are no rewards or penalties in other cases. Users can adjust the difficulty by modifying the monster’s HP and the number of agents.

\textbf{\textit{Flag Capture}} - Scenario designed for multi-team gaming tasks. It involves several robot teams gaming to capture a flag. The closest robot can pick up the flag, and their teammates must defend it from other teams. At the end of each episode, the team that held the flag the longest wins. The number and team affiliations of agents can be freely changed to modify the features and difficulty of the tasks.

\textbf{\textit{Navigation Game}} - Scenario designed for heterogeneous and two-team zero-sum gaming tasks. It includes two landmarks, a keeper team, and a navigator team. Although they cannot attack each other, the ground keeper can drive away the air navigator, while the ground navigator can drive away the ground keeper. If the air navigator stays over any landmark for a certain period, the navigator team is deemed to have won. The rewards for the two teams are zero-sum. The number of agents in each team can be freely changed, altering the difficulty of the tasks.

\section{Experiments}
\label{sec:Experiments}

\begin{table*}[ht]
\caption{Description of example tasks in the experiments.}
\label{tab:task-table}
\vskip 0.15in
\centering
{\tiny 
\begin{tabular}{lcccc}
\toprule
\textbf{Example Task} & \textbf{MARL Agents} & \textbf{Other Entities} & \textbf{Features} & \textbf{Remark} \\
\midrule

\textit{metal\_clash\_5lc\_5mc} & 5 Laser-Car, 5 Missile-Car & 5 Laser-Car, 5 Missile-Car & Heterogeneous & - \\

\textit{metal\_clash\_het\_10} & 4 Laser-Car, 4 Missile-Car, 2 Support-Drone & 4 Laser-Car, 4 Missile-Car, 2 Support-Drone & Heterogeneous & -\\

\textit{metal\_clash\_het\_8\_vs\_10} & 4 Laser-Car, 2 Missile-Car, 2 Support-Drone & 4 Laser-Car, 4 Missile-Car, 2 Support-Drone & Heterogeneous & -\\

\textit{metal\_clash\_hom\_50} & 50 Laser-Car & 50 Laser-Car & Large-Scale & -\\

\textit{metal\_clash\_het\_50} & 20 Laser-Car, 20 Missile-Car, 10 Support-Drone & 20 Laser-Car, 20 Missile-Car, 10 Support-Drone & Heterogeneous + Large-Scale & -\\

\textit{metal\_clash\_het\_100} & 40 Laser-Car, 40 Missile-Car, 20 Support-Drone & 40 Laser-Car, 40 Missile-Car, 20 Support-Drone & Heterogeneous + Large-Scale & -\\

\textit{monster\_crisis\_easy} & 8 Mushroom & 1 Monster & Sparse Team Reward & HP of Monster is 400\\

\textit{monster\_crisis\_medium} & 8 Mushroom & 1 Monster & Sparse Team Reward & HP of Monster is 600\\

\textit{monster\_crisis\_hard} & 8 Mushroom & 1 Monster & Sparse Team Reward & HP of Monster is 800\\

\textit{flag\_capture\_1script} & 15 Robot & 15 Robot & Zero-Sum Game & - \\

\textit{flag\_capture\_2scripts} & 15 Robot & 2 * [15 Robot] & Multi-Team & -\\

\textit{flag\_capture\_2scripts\_hard} & 10 Robot & 2 * [15 Robot] & Multi-Team & -\\

\textit{navigation\_game\_5\_vs\_2} & 3 Ground-Navigator, 2 Air-Navigator & 2 Ground-Keeper & Heterogeneous + Game & -\\ 

\textit{navigation\_game\_4\_vs\_2} & 2 Ground-Navigator, 2 Air-Navigator & 2 Ground-Keeper & Heterogeneous + Game & -\\ 

\textit{navigation\_game\_3\_vs\_2} & 2 Ground-Navigator, 1 Air-Navigator & 2 Ground-Keeper & Heterogeneous + Game & -\\

\bottomrule
\end{tabular}
}
\vskip -0.1in
\end{table*}

\begin{figure*}[!t]
    \centering
    \includegraphics[width=0.95\textwidth]
    {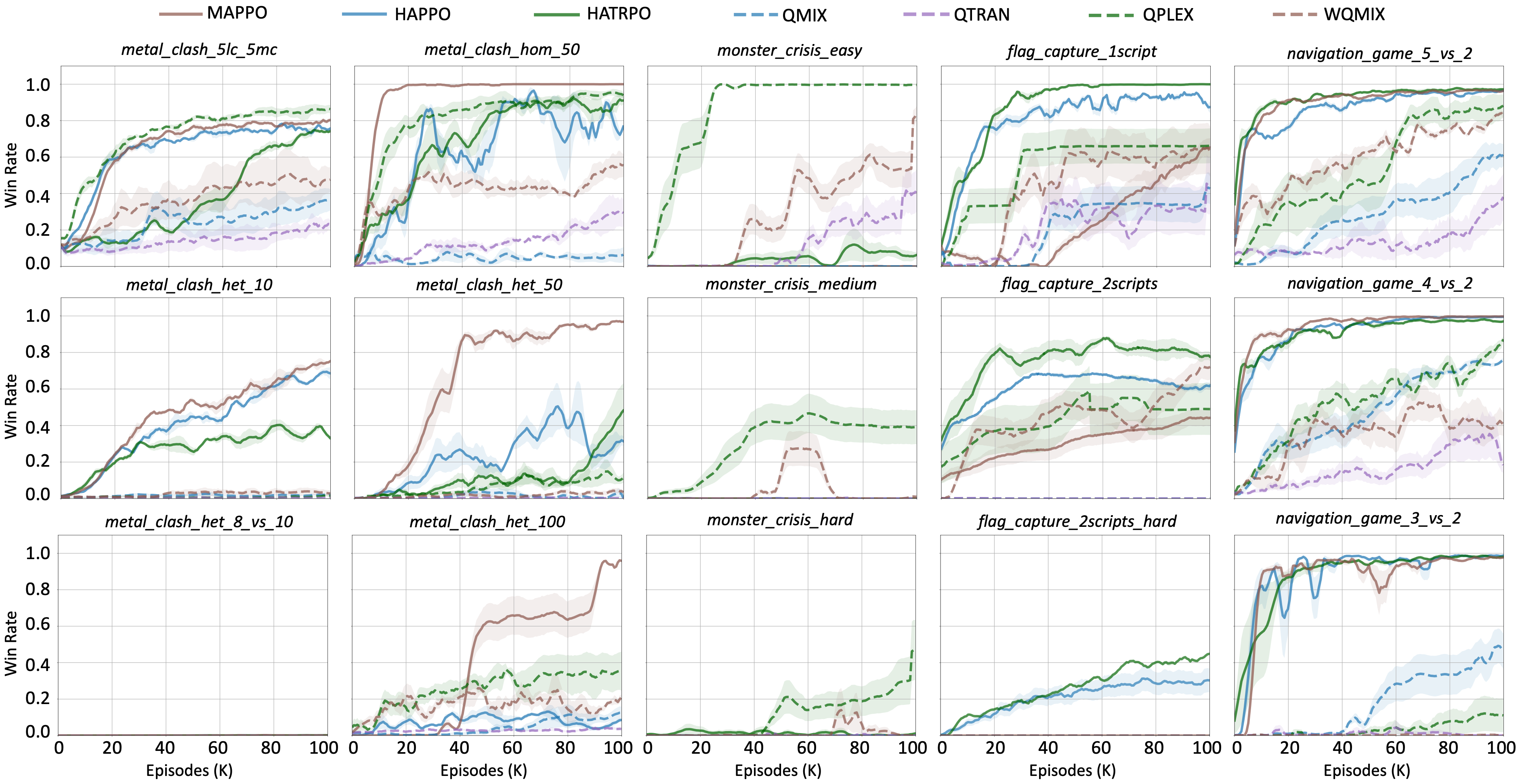}
    \caption{The comparison of test win rate for all tested algorithms across 15 tasks. The shadowed area depicts the 95\% confidence interval.}
    \label{fig:15Result}
\end{figure*} 

To demonstrate the utility of Unreal-MAP, we develop 15 example tasks and deploy several MARL algorithms across them. We find that by altering the properties of scenarios, it is capable of developing tasks that challenge current algorithms. Additionally, various algorithms exhibit superior performance in their areas of strength. It is worth mentioning that we DO NOT intend to develop the example tasks as benchmarks, but rather use them to show that based on Unreal-MAP, it is possible to develop extensible tasks that facilitate the deployment of popular MARL algorithms.

We then evaluate the training efficiency and resource consumption of Unreal-MAP. We find that changing the speed within a process only affects the CPU utilization of the device, and it is possible to use more CPU resources under memory-limited conditions by adjusting the time dilation factor, or vice versa. We also discover that simulation tasks for a scale of 20 agents developed based on Unreal-MAP, can achieve physical simulation frames at the 1M level per second, enabling the corresponding training tasks to be completed within hours.

Finally, we implement a sim2real demo based on the navigation-game scenario, to demonstrate the potential of Unreal-MAP in \textit{simulating real-world environments} and \textit{deploying algorithms in the real world}.

\begin{figure}[!h] 
    \centering
    \includegraphics[width=1.0\linewidth]
    {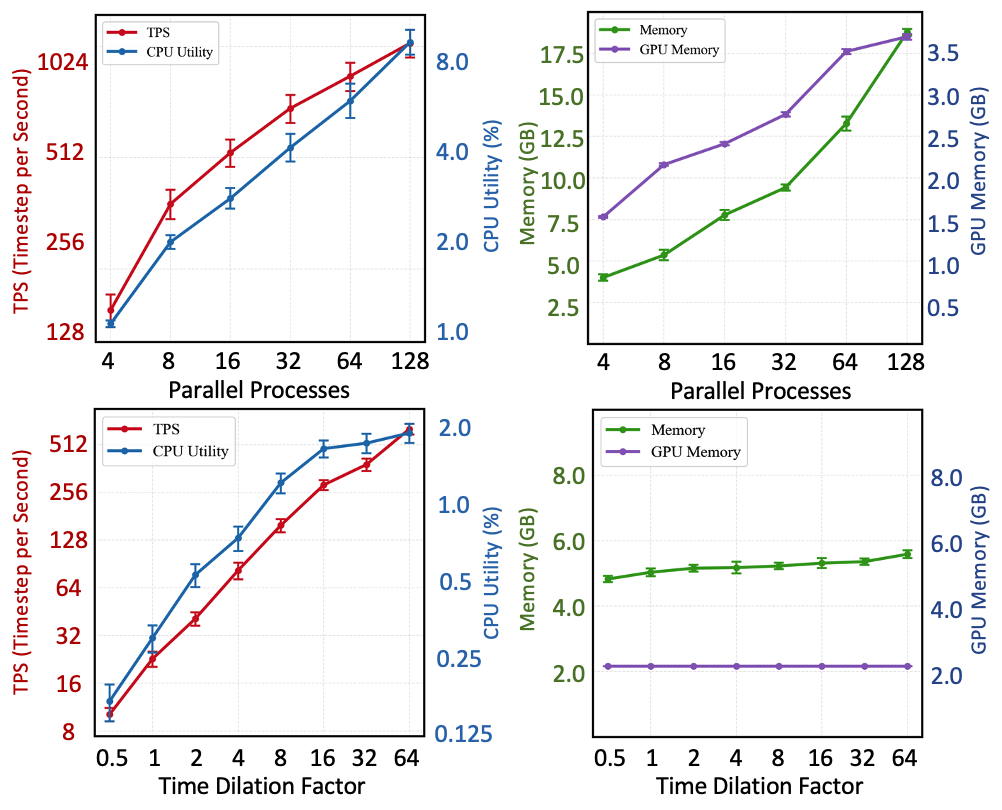}
    \caption{The impact of the number of parallel processes and the time dilation factor in Unreal-MAP on simulation efficiency and computational resource consumption.}
  
    \label{fig:Eff}
\end{figure}

\subsection{Performance in Example Tasks}

We develop 15 example tasks based on 4 scenarios from Unreal-MAP, as detailed in Table~\ref{tab:task-table}. Based on HMAP, we deploy 7 SOTA MARL algorithms on all tasks, including the actor-critic-based algorithms as  MAPPO~\citep{MAPPO}, HATRPO and HAPPO~\citep{HARL}, as well as the value-based algorithms as QMIX~\citep{QMIX}, QTRAN~\citep{QTRAN}, QPLEX~\citep{QPLEX} and WQMIX~\citep{WQMIX}.
To ensure a fair comparison, the main network of each algorithm is preserved uniform, and hyperparameters are standardized (refer to Appendix~\ref{sec:Hyperparameter Details} for details).
The effectiveness of the training is tested after every 1280 episodes. The average win rates are calculated based on 512 episodes per test, across 5 or more random seeds. The results are illustrated in Figure~\ref{fig:15Result}, where the lines represent the mean values and the shadowed areas indicate the 95\% confidence interval.

The results from Figure~\ref{fig:15Result} show that Unreal-MAP is capable of developing MARL-compatible multi-agent tasks. By comparing the test results of different tasks developed from the same scenario, it is evident that changing 1) the property of individual entities or agents 2) the types of agents 3) the number of agents 4) the number of agent teams can effectively alter the features and difficulty of the tasks. 

By comparing the results of different algorithms, we also discover some interesting findings. Value-based algorithms significantly outperform actor-critic(AC)-based methods in tasks with sparse team rewards, but not in other tasks compared to AC's SOTA algorithms. This is because value-based algorithms, which focus on value decomposition, are better at solving hard-to-decompose team reward problems. Within AC algorithms, MAPPO performs better in large-scale tasks due to synchronous updates and parameter sharing, while HAPPO and HATRPO perform better in multi-team (unstable environmental changes) tasks due to asynchronous monotonic updates. Additional experiments and analysis details can be found in Appendix~\ref{app:Additional Experiments and Analysis}.


\begin{figure*}[!t]
    \centering
    \includegraphics[width=0.95\textwidth]
    {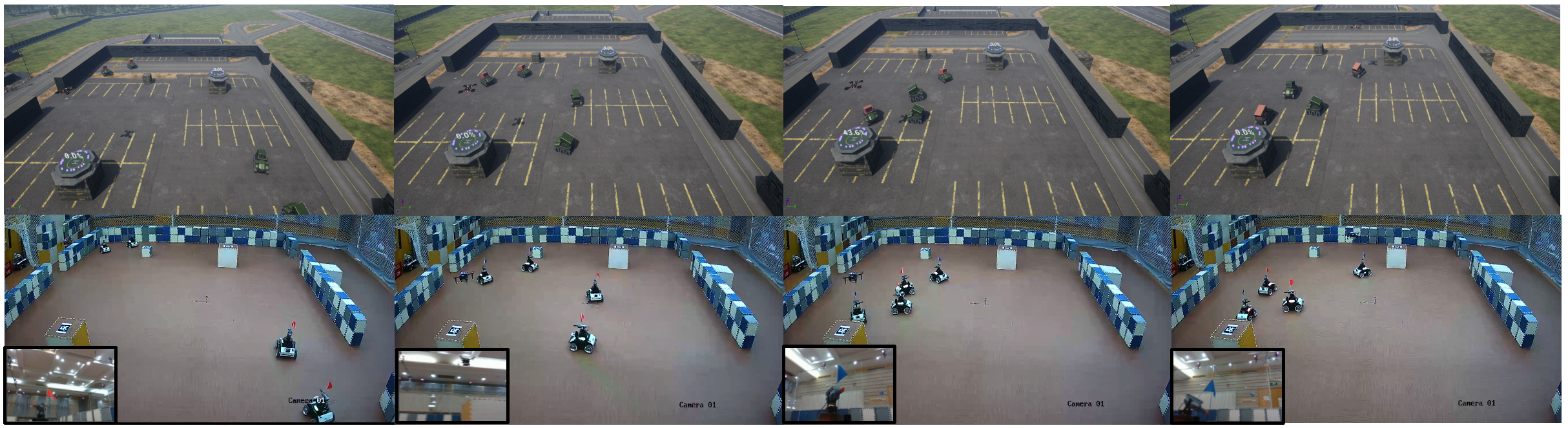}
    \caption{Snapshots from Unreal-MAP-simulated and real-world scenarios. 
    The top four subfigures shows snapshots of multiple agents deploying well-trained policies only in virtual scenarios. The bottom four subfigures shows deployed policies in real-world scenarios at the same timesteps.}
    \label{fig:sim2real}
\end{figure*}

\subsection{Efficiency and Computational Consumption}
\label{sec:Eff}
We conduct experiments on the efficiency index and resource consumption indices of Unreal-MAP. The efficiency index adopted is TPS, i.e., the number of \textit{virtual Timesteps} Per \textit{real Second}. The resource consumption indices include CPU utilization, memory occupancy, and GPU memory occupancy.
Unreal-MAP has two dimensions to control simulation efficiency and resource consumption: the number of processes and the speed within each process. Hence, we conduct related experiments, obtaining the relationship curves between the above two dimensions and four indices, as shown in Figure~\ref{fig:Eff}. All experiments are conducted on a Linux server equipped with 8 NVIDIA RTX3090 GPUs, and the tested task is \textit{metal\_clash\_5lc\_5mc}, details of experiments are available in Appendix~\ref{app:Eff}.

Through the experimental results, we can see that increasing the number of parallel processes increases the TPS at the cost of increasing all resource consumption indices. However, \textbf{increasing the time dilation factor only increases CPU utilization, with almost no effect on memory and GPU memory.} The time dilation factor is roughly proportional to TPS and CPU utilization. This means that under limited memory resources, training efficiency can be improved by increasing the time dilation factor to fully utilize the CPU; similarly, under limited CPU resources, reducing the time dilation factor and increasing the number of processes can avoid the waste of computing resources.

Furthermore, according to our experimental results, when the number of processes is 8 and the time dilation factor is 32, TPS can reach 400. Since the maximum number of episode steps is 100, training 1024 episodes on the \textit{metal\_clash\_5lc\_5mc} task takes less than 2 minutes. This means that under such settings, \textbf{the tested server can simultaneously support 50 such tasks (each with 20 agents) and complete all training tasks (100k episodes) within 3 hours.} When the number of processes reaches 128 and the time dilation factor is set to 32, the TPS can reach 1000+, and the training task can be completed in about an hour.

It is important to emphasize that TPS here counts for the number of virtual Unreal-MAP timesteps per real second. Since this is a simulation of 20 agents, and each timestep in Unreal-MAP undergoes 1280 frames of calculations for environmental dynamics and kinematics to maintain fine state transitions (details in Appendix~\ref{app:Additional Experiments and Analysis}), this means that \textbf{the speed of simulation physics frame calculation can reach the 1M level}, which is a highly efficient computation.

\subsection{Physical Experiment}
\label{sec:Physical Experiment}

We conduct this experiment to demonstrate the potential of Unreal-MAP in bridging the sim-to-real gap. Firstly, we construct a real-world experimental setup, which consists of a motion capture system, a communication system, several autonomous UGVs and UAVs, and a number of physical entities. Subsequently, we develop the \textit{landmark\_conquer} scenario through Unreal-MAP, wherein the entities are proportionally replicated from the physical setup, and the kinematics of the unmanned units are also recreated. Ultimately, we develop an \textit{algorithm-Unreal-MAP-hardware}
framework, with details presented in Appendix~\ref{sec:Physical Experiment Details}.

During the training phase, the algorithmic side, represented by HMAP, interact with Unreal-MAP to train policies within the simulated scenarios. In the execution phase, the physical system relay global information captured by the motion capture system and first-person view data from the vehicles' cameras to Unreal-MAP. Unreal-MAP then update its internal environment with this information and transmit the filtered observational data to HMAP. The algorithm within HMAP generate action commands based on these observations, which are conveyed to Unreal-MAP. Unreal-MAP execute virtual state transitions based on these commands, and concurrently transmit the decomposed action information to the real-world setup for execution by the autonomous vehicles/drones.

Figure~\ref{fig:sim2real} presents snapshots from both the virtual and the real-world scenarios. The experimental results indicate that the whole system can successfully replicate the policies of the multi-agent system from the virtual environment within the physical setup.


\section{Limitations and Future Work}

In this paper, we propose Unreal-MAP, an MARL general platform based on the Unreal Engine. We demonstrate the effects of deploying sample tasks developed on this platform and cutting-edge MARL algorithms. However, Unreal-MAP is not perfect. One limitation is that Unreal-MAP still requires the CPU to handle scene logic, physical calculations, network communications, and other tasks, and its training rate cannot match some purely-GPU-implemented environments. Additionally, although Unreal-MAP has made all development tasks achievable solely through Python and Blueprint, the MARL community may not yet be accustomed to Blueprint programming.

We plan to further develop Unreal-MAP in two directions. The first is related to generative AI, where we plan to first integrate large models at the Python-based interface layer to assist users in quickly customizing MARL tasks. The second is focused on sim2real. We will develop a complete, plug-and-play sim2real toolkit based on UMAP, mapping real-world demands into the virtual world of Unreal-MAP, thereby pushing the practical application of MARL to the next level.

\section*{Impact Statement}

The further development of MARL requires simulation environments with high scalability and realistic physical modelling capabilities. By combining MARL with game engines, our work will help develop a general platform and cater to the development trend of generative AI. Our work also directly bridges the MARL community and the game development community, with great potential to further enhance the development of the MARL and RL fields. However, directly adopting content from the game engine community is double-edged, and we need to ensure that it does not bring potential negative impacts. To this end, we have added a discussion on ethical review content in Appendix~\ref{Ethics}.




\nocite{langley00}

\bibliography{example_paper}

\begin{thebibliography}{47}
\providecommand{\natexlab}[1]{#1}
\providecommand{\url}[1]{\texttt{#1}}
\expandafter\ifx\csname urlstyle\endcsname\relax
  \providecommand{\doi}[1]{doi: #1}\else
  \providecommand{\doi}{doi: \begingroup \urlstyle{rm}\Url}\fi

\bibitem[Authors(2024)]{Genesis}
Authors, G.
\newblock Genesis: A universal and generative physics engine for robotics and
  beyond, December 2024.
\newblock URL \url{https://github.com/Genesis-Embodied-AI/Genesis}.

\bibitem[Baker et~al.(2019)Baker, Kanitscheider, Markov, Wu, Powell, McGrew,
  and Mordatch]{Hide-and-Seek}
Baker, B., Kanitscheider, I., Markov, T., Wu, Y., Powell, G., McGrew, B., and
  Mordatch, I.
\newblock Emergent tool use from multi-agent autocurricula.
\newblock \emph{arXiv preprint arXiv:1909.07528}, 2019.

\bibitem[Bala et~al.(2024)Bala, Cui, Ding, Ge, Hao, Hasselgren, Huffman, Jin,
  Lewis, Li, et~al.]{Edify_3D}
Bala, M., Cui, Y., Ding, Y., Ge, Y., Hao, Z., Hasselgren, J., Huffman, J., Jin,
  J., Lewis, J., Li, Z., et~al.
\newblock Edify 3d: Scalable high-quality 3d asset generation.
\newblock \emph{arXiv preprint arXiv:2411.07135}, 2024.

\bibitem[Bard et~al.(2020)Bard, Foerster, Chandar, Burch, Lanctot, Song,
  Parisotto, Dumoulin, Moitra, Hughes, et~al.]{Hanabi}
Bard, N., Foerster, J.~N., Chandar, S., Burch, N., Lanctot, M., Song, H.~F.,
  Parisotto, E., Dumoulin, V., Moitra, S., Hughes, E., et~al.
\newblock The hanabi challenge: A new frontier for ai research.
\newblock \emph{Artificial Intelligence}, 280:\penalty0 103216, 2020.

\bibitem[Booth \& Booth(2019)Booth and Booth]{Marathon}
Booth, J. and Booth, J.
\newblock Marathon environments: Multi-agent continuous control benchmarks in a
  modern video game engine.
\newblock \emph{arXiv preprint arXiv:1902.09097}, 2019.

\bibitem[Boyd(2017)]{boyd2017implementing}
Boyd, R.
\newblock \emph{Implementing reinforcement learning in unreal engine 4 with
  blueprint}.
\newblock PhD thesis, University Honors College, Middle Tennessee State
  University, 2017.

\bibitem[Boyd \& Barbosa(2017)Boyd and Barbosa]{boyd2017reinforcement}
Boyd, R.~A. and Barbosa, S.~E.
\newblock Reinforcement learning for all: An implementation using unreal engine
  blueprint.
\newblock In \emph{2017 International Conference on Computational Science and
  Computational Intelligence (CSCI)}, pp.\  787--792. IEEE, 2017.

\bibitem[Brockman(2016)]{Gym}
Brockman, G.
\newblock Openai gym.
\newblock \emph{arXiv preprint arXiv:1606.01540}, 2016.

\bibitem[Chen et~al.(2020)Chen, Chang, and Zhang]{chen2020autonomous}
Chen, Y.-J., Chang, D.-K., and Zhang, C.
\newblock Autonomous tracking using a swarm of uavs: A constrained multi-agent
  reinforcement learning approach.
\newblock \emph{IEEE Transactions on Vehicular Technology}, 69\penalty0
  (11):\penalty0 13702--13717, 2020.

\bibitem[Ellis et~al.(2022)Ellis, Moalla, Samvelyan, Sun, Mahajan, Foerster,
  and Whiteson]{SMACv2}
Ellis, B., Moalla, S., Samvelyan, M., Sun, M., Mahajan, A., Foerster, J.~N.,
  and Whiteson, S.
\newblock Smacv2: An improved benchmark for cooperative multi-agent
  reinforcement learning.
\newblock \emph{arXiv preprint arXiv:2212.07489}, 2022.

\bibitem[Fu et~al.(2024)Fu, Pu, Pan, Qiu, and Yi]{Fuzzy}
Fu, Q., Pu, Z., Pan, Y., Qiu, T., and Yi, J.
\newblock Fuzzy feedback multi-agent reinforcement learning for adversarial
  dynamic multi-team competitions.
\newblock \emph{IEEE Transactions on Fuzzy Systems}, 2024.

\bibitem[Gronauer \& Diepold(2022)Gronauer and Diepold]{POMG_survey}
Gronauer, S. and Diepold, K.
\newblock Multi‑agent deep reinforcement learning: a survey.
\newblock \emph{Artificial Intelligence Review}, 55:\penalty0 895--943, 2022.

\bibitem[Haarnoja et~al.(2018)Haarnoja, Zhou, Abbeel, and Levine]{SAC}
Haarnoja, T., Zhou, A., Abbeel, P., and Levine, S.
\newblock Soft actor-critic: Off-policy maximum entropy deep reinforcement
  learning with a stochastic actor.
\newblock In \emph{International conference on machine learning}, pp.\
  1861--1870. PMLR, 2018.

\bibitem[Hu et~al.(2021)Hu, Jiang, Harding, Wu, and Liao]{pymarl2}
Hu, J., Jiang, S., Harding, S.~A., Wu, H., and Liao, S.-w.
\newblock Rethinking the implementation tricks and monotonicity constraint in
  cooperative multi-agent reinforcement learning.
\newblock \emph{arXiv preprint arXiv:2102.03479}, 2021.

\bibitem[Juliani(2018)]{Unity}
Juliani, A.
\newblock Unity: A general platform for intelligent agents.
\newblock \emph{arXiv preprint arXiv:1809.02627}, 2018.

\bibitem[Kaffee et~al.(2023)Kaffee, Arora, Talat, and Augenstein]{ThornyRoses}
Kaffee, L.-A., Arora, A., Talat, Z., and Augenstein, I.
\newblock Thorny roses: Investigating the dual use dilemma in natural language
  processing.
\newblock \emph{arXiv preprint arXiv:2304.08315}, 2023.

\bibitem[Kalashnikov et~al.(2018)Kalashnikov, Irpan, Pastor, Ibarz, Herzog,
  Jang, Quillen, Holly, Kalakrishnan, Vanhoucke,
  et~al.]{kalashnikov2018scalable}
Kalashnikov, D., Irpan, A., Pastor, P., Ibarz, J., Herzog, A., Jang, E.,
  Quillen, D., Holly, E., Kalakrishnan, M., Vanhoucke, V., et~al.
\newblock Scalable deep reinforcement learning for vision-based robotic
  manipulation.
\newblock In \emph{Conference on robot learning}, pp.\  651--673. PMLR, 2018.

\bibitem[Kaup et~al.(2024)Kaup, Wolff, Hwang, Mayer, and Bruni]{kaup2024review}
Kaup, M., Wolff, C., Hwang, H., Mayer, J., and Bruni, E.
\newblock A review of nine physics engines for reinforcement learning research.
\newblock \emph{arXiv preprint arXiv:2407.08590}, 2024.

\bibitem[Kurach et~al.(2020)]{GRF}
Kurach et~al.
\newblock Google research football: A novel reinforcement learning environment.
\newblock In \emph{Proceedings of the AAAI conference on artificial
  intelligence}, volume~34, pp.\  4501--4510, 2020.

\bibitem[Littman(1994)]{Markov_game}
Littman, M.~L.
\newblock Markov games as a framework for multi-agent reinforcement learning.
\newblock In \emph{Machine learning proceedings 1994}, pp.\  157--163.
  Elsevier, 1994.

\bibitem[Liu et~al.(2023)Liu, Shao, Chen, Qu, Wang, Ye, Tu, Qin, Feng, Lai,
  et~al.]{HoK3v3}
Liu, L., Shao, J., Chen, X., Qu, Y., Wang, B., Ye, Z., Tu, Y., Qin, H., Feng,
  Y.~J., Lai, L., et~al.
\newblock Hok3v3: an environment for generalization in heterogeneous
  multi-agent reinforcement learning.
\newblock 2023.

\bibitem[Lu et~al.(2024)Lu, Shu, Xiao, Ye, Wang, Peng, Wei, Khashabi,
  Chellappa, Yuille, et~al.]{lu2024genex}
Lu, T., Shu, T., Xiao, J., Ye, L., Wang, J., Peng, C., Wei, C., Khashabi, D.,
  Chellappa, R., Yuille, A., et~al.
\newblock Genex: Generating an explorable world.
\newblock \emph{arXiv preprint arXiv:2412.09624}, 2024.

\bibitem[Mnih et~al.(2015)Mnih, Kavukcuoglu, Silver, Rusu, Veness, Bellemare,
  Graves, Riedmiller, Fidjeland, Ostrovski, et~al.]{DQN}
Mnih, V., Kavukcuoglu, K., Silver, D., Rusu, A.~A., Veness, J., Bellemare,
  M.~G., Graves, A., Riedmiller, M., Fidjeland, A.~K., Ostrovski, G., et~al.
\newblock Human-level control through deep reinforcement learning.
\newblock \emph{nature}, 518\penalty0 (7540):\penalty0 529--533, 2015.

\bibitem[Mordatch \& Abbeel(2017)Mordatch and Abbeel]{MPE}
Mordatch, I. and Abbeel, P.
\newblock Emergence of grounded compositional language in multi-agent
  populations.
\newblock \emph{arXiv preprint arXiv:1703.04908}, 2017.

\bibitem[NVIDIA(2024)]{nvidia_ace}
NVIDIA.
\newblock Ace, 2024.
\newblock URL \url{https://developer.nvidia.com/ace}.

\bibitem[Oroojlooy \& Hajinezhad(2023)Oroojlooy and Hajinezhad]{2023review}
Oroojlooy, A. and Hajinezhad, D.
\newblock A review of cooperative multi-agent deep reinforcement learning.
\newblock \emph{Applied Intelligence}, 53\penalty0 (11):\penalty0 13677--13722,
  2023.

\bibitem[Peng et~al.(2021{\natexlab{a}})Peng, Rashid, Schroeder~de Witt,
  Kamienny, Torr, B{\"o}hmer, and Whiteson]{MAMuJoCo}
Peng, B., Rashid, T., Schroeder~de Witt, C., Kamienny, P.-A., Torr, P.,
  B{\"o}hmer, W., and Whiteson, S.
\newblock Facmac: Factored multi-agent centralised policy gradients.
\newblock \emph{Advances in Neural Information Processing Systems},
  34:\penalty0 12208--12221, 2021{\natexlab{a}}.

\bibitem[Peng et~al.(2021{\natexlab{b}})Peng, Li, Hui, Liu, and
  Zhou]{autonomous_vehicles_1}
Peng, Z., Li, Q., Hui, K.~M., Liu, C., and Zhou, B.
\newblock Learning to simulate self-driven particles system with coordinated
  policy optimization.
\newblock volume~34, pp.\  10784--10797, 2021{\natexlab{b}}.

\bibitem[Rashid et~al.(2020{\natexlab{a}})Rashid, Farquhar, Peng, and
  Whiteson]{WQMIX}
Rashid, T., Farquhar, G., Peng, B., and Whiteson, S.
\newblock Weighted qmix: Expanding monotonic value function factorisation for
  deep multi-agent reinforcement learning.
\newblock \emph{Advances in neural information processing systems},
  33:\penalty0 10199--10210, 2020{\natexlab{a}}.

\bibitem[Rashid et~al.(2020{\natexlab{b}})Rashid, Samvelyan, De~Witt, Farquhar,
  Foerster, and Whiteson]{QMIX}
Rashid, T., Samvelyan, M., De~Witt, C.~S., Farquhar, G., Foerster, J., and
  Whiteson, S.
\newblock Monotonic value function factorisation for deep multi-agent
  reinforcement learning.
\newblock \emph{The Journal of Machine Learning Research}, 21\penalty0
  (1):\penalty0 7234--7284, 2020{\natexlab{b}}.

\bibitem[Rutherford et~al.(2024)Rutherford, Ellis, Gallici, Cook, Lupu,
  Ingvarsson, Willi, Hammond, Khan, de~Witt, et~al.]{JaxMARL}
Rutherford, A., Ellis, B., Gallici, M., Cook, J., Lupu, A., Ingvarsson, G.,
  Willi, T., Hammond, R., Khan, A., de~Witt, C.~S., et~al.
\newblock Jaxmarl: Multi-agent rl environments and algorithms in jax.
\newblock In \emph{The Thirty-eight Conference on Neural Information Processing
  Systems Datasets and Benchmarks Track}, 2024.

\bibitem[Samvelyan et~al.(2019)Samvelyan, Rashid, de~Witt, Farquhar, Nardelli,
  Rudner, Hung, Torr, Foerster, and Whiteson]{SMAC}
Samvelyan, M., Rashid, T., de~Witt, C.~S., Farquhar, G., Nardelli, N., Rudner,
  T. G.~J., Hung, C.-M., Torr, P. H.~S., Foerster, J., and Whiteson, S.
\newblock {The} {StarCraft} {Multi}-{Agent} {Challenge}.
\newblock \emph{CoRR}, abs/1902.04043, 2019.

\bibitem[Sapio \& Ratini(2022)Sapio and Ratini]{URLT}
Sapio, F. and Ratini, R.
\newblock Developing and testing a new reinforcement learning toolkit with
  unreal engine.
\newblock In \emph{International Conference on Human-Computer Interaction},
  pp.\  317--334. Springer, 2022.

\bibitem[Schuiling(2017)]{GameplayFootball}
Schuiling, B.~K.
\newblock Gameplayfootball.
\newblock \url{https://github.com/BazkieBumpercar/GameplayFootball/}, 2017.

\bibitem[Son et~al.(2019)Son, Kim, Kang, Hostallero, and Yi]{QTRAN}
Son, K., Kim, D., Kang, W.~J., Hostallero, D.~E., and Yi, Y.
\newblock Qtran: Learning to factorize with transformation for cooperative
  multi-agent reinforcement learning.
\newblock In \emph{International conference on machine learning}, pp.\
  5887--5896. PMLR, 2019.

\bibitem[Suarez et~al.(2021)Suarez, Du, Zhu, Mordatch, and Isola]{NerualMMO}
Suarez, J., Du, Y., Zhu, C., Mordatch, I., and Isola, P.
\newblock The neural mmo platform for massively multiagent research.
\newblock \emph{arXiv preprint arXiv:2110.07594}, 2021.

\bibitem[Templet(2021)]{templet2021game}
Templet, R.~M.
\newblock \emph{Game Physics: An Analysis of Physics Engines for First-Time
  Physics Developers}.
\newblock PhD thesis, California State University, Northridge, 2021.

\bibitem[Terry et~al.(2021)Terry, Black, Grammel, Jayakumar, Hari, Sullivan,
  Santos, Dieffendahl, Horsch, Perez-Vicente, et~al.]{Pettingzoo}
Terry, J., Black, B., Grammel, N., Jayakumar, M., Hari, A., Sullivan, R.,
  Santos, L.~S., Dieffendahl, C., Horsch, C., Perez-Vicente, R., et~al.
\newblock Pettingzoo: Gym for multi-agent reinforcement learning.
\newblock \emph{Advances in Neural Information Processing Systems},
  34:\penalty0 15032--15043, 2021.

\bibitem[Todorov et~al.(2012)Todorov, Erez, and Tassa]{Mujoco}
Todorov, E., Erez, T., and Tassa, Y.
\newblock Mujoco: A physics engine for model-based control.
\newblock In \emph{2012 IEEE/RSJ international conference on intelligent robots
  and systems}, pp.\  5026--5033. IEEE, 2012.

\bibitem[Vinyals et~al.(2019)Vinyals, Babuschkin, Czarnecki, Mathieu, Dudzik,
  Chung, Choi, Powell, Ewalds, Georgiev, et~al.]{vinyals2019grandmaster}
Vinyals, O., Babuschkin, I., Czarnecki, W.~M., Mathieu, M., Dudzik, A., Chung,
  J., Choi, D.~H., Powell, R., Ewalds, T., Georgiev, P., et~al.
\newblock Grandmaster level in starcraft ii using multi-agent reinforcement
  learning.
\newblock \emph{Nature}, 575\penalty0 (7782):\penalty0 350--354, 2019.

\bibitem[Vohera et~al.(2021)Vohera, Chheda, Chouhan, Desai, and
  Jain]{vohera2021game}
Vohera, C., Chheda, H., Chouhan, D., Desai, A., and Jain, V.
\newblock Game engine architecture and comparative study of different game
  engines.
\newblock In \emph{2021 12th International Conference on Computing
  Communication and Networking Technologies (ICCCNT)}, pp.\  1--6. IEEE, 2021.

\bibitem[Wang et~al.(2020)Wang, Ren, Liu, Yu, and Zhang]{QPLEX}
Wang, J., Ren, Z., Liu, T., Yu, Y., and Zhang, C.
\newblock Qplex: Duplex dueling multi-agent q-learning.
\newblock \emph{arXiv preprint arXiv:2008.01062}, 2020.

\bibitem[Wheeler(2023)]{wheeler2023reinforcement}
Wheeler, J.~B.
\newblock Reinforcement learning framework for the unreal engine.
\newblock 2023.

\bibitem[Yu et~al.(2022)Yu, Velu, Vinitsky, Gao, Wang, Bayen, and Wu]{MAPPO}
Yu, C., Velu, A., Vinitsky, E., Gao, J., Wang, Y., Bayen, A., and Wu, Y.
\newblock The surprising effectiveness of ppo in cooperative multi-agent games.
\newblock \emph{Advances in Neural Information Processing Systems},
  35:\penalty0 24611--24624, 2022.

\bibitem[Zhang et~al.(2022)Zhang, Zhang, Yang, Chen, Zheng, Yang, Li, Zhou,
  Niu, and Liu]{GoBigger}
Zhang, M., Zhang, S., Yang, Z., Chen, L., Zheng, J., Yang, C., Li, C., Zhou,
  H., Niu, Y., and Liu, Y.
\newblock Gobigger: A scalable platform for cooperative-competitive multi-agent
  interactive simulation.
\newblock In \emph{The Eleventh International Conference on Learning
  Representations}, 2022.

\bibitem[Zheng et~al.(2018)Zheng, Yang, Cai, Zhou, Zhang, Wang, and Yu]{MAgent}
Zheng, L., Yang, J., Cai, H., Zhou, M., Zhang, W., Wang, J., and Yu, Y.
\newblock Magent: A many-agent reinforcement learning platform for artificial
  collective intelligence.
\newblock In \emph{Proceedings of the AAAI conference on artificial
  intelligence}, volume~32, 2018.

\bibitem[Zhong et~al.(2024)Zhong, Kuba, Feng, Hu, Ji, and Yang]{HARL}
Zhong, Y., Kuba, J.~G., Feng, X., Hu, S., Ji, J., and Yang, Y.
\newblock Heterogeneous-agent reinforcement learning.
\newblock \emph{Journal of Machine Learning Research}, 25\penalty0
  (1-67):\penalty0 1, 2024.

\end{thebibliography}
\bibliographystyle{icml2025}

\newpage
\appendix
\onecolumn





\newpage
\section{Open Source Statement}
\label{App:Open Source Statement}

We list the contents included in our open-source project, which not only encompasses Unreal-MAP and HMAP but also the corresponding ecosystem, including tutorials, Docker environments, etc. The open-source project is as follows:

\begin{enumerate}
    \item \textbf{Regarding code environment configuration:} We release a Docker image supporting Unreal-MAP and HMAP services on Docker Hub. This image includes the HMAP framework, a default version of Unreal-MAP's compiled binary files, and a series of environment configurations.
    \item \textbf{Regarding Unreal-MAP:} We publish Unreal-MAP's usage tutorials and one-click deployment scripts on GitHub. These scripts facilitate the compilation of rendering/training-only binary files for various platforms and automate the downloading of large files. The Unreal project and the modified Unreal Engine of Unreal-MAP will be available on a cloud drive, accessible for automatic download via Python scripts.
    \item \textbf{Regarding HMAP:} We publish HMAP's usage tutorials and its entire content to GitHub. This content includes the core of HMAP, wrappers for all supported environments, built-in algorithms, and algorithms from third-party frameworks.
    \item \textbf{Future Plans for Open Source Work:} We will continue to maintain all GitHub repositories, develop new scenarios, incorporate more algorithms from third-party frameworks, and develop sim-to-real related toolkits.
\end{enumerate}

\section{Related Work}
\label{Sec:related work and comparison}

\begin{table}[ht]
\centering
\caption{Comparison of Unreal-MAP with other related MARL simulation environments.}
\label{Tab: Comparison}
\label{tab:example}
{\tiny 
\begin{tabular}{cccccccc}
\toprule

& \textbf{Unreal-MAP(Ours)} & MPE & MAgent & Hanabi & NeuralMMO  & GoBigger & JaxMARL\\ \midrule

Heterogenous Support  
& \ding{51} & \ding{51}   & \ding{51}   & \textbf{--} &\ding{51}    & \textbf{--} & \ding{51}   \\
Large-Scale Support   
& \ding{51} & \textbf{--} & \ding{51}   & \textbf{--} & \ding{51}   & \ding{51}   & \ding{51}  \\
Multi-Team Support          
& \ding{51} & \textbf{--} & \textbf{--} &\textbf{--}  & \textbf{--} & \ding{51}   & \textbf{--}  \\
Mixed-Game Support
& \ding{51} & \ding{51}   & \ding{51}   & \textbf{--} & \ding{51}   & \ding{51}   & \ding{51}  \\

3D Physics Engine              
& \ding{51} & \textbf{--} & \textbf{--} & \textbf{--} & \textbf{--} & \textbf{--} & \textbf{--} \\

Fully Open Source        
& \ding{51} & \ding{51} & \ding{51} & \ding{51} & \textbf{--} & \textbf{--} & \ding{51}  \\

All Elements Customizable
& \ding{51} & \ding{51} & \ding{51} & \ding{51} & \textbf{--} & \textbf{--} & \ding{51}  \\

Controllable Time-flow Speed 
& \ding{51} & \textbf{--} & \textbf{--} & \textbf{--} & \textbf{--} & \textbf{--} & \textbf{--} \\

Rendering Training
& \ding{51} & \textbf{--} & \textbf{--} & \textbf{--}& \ding{51} & \textbf{--} & \textbf{--}  \\
 \midrule

& GRF & SMAC & SMACv2 & Hide-and-Seek & HoK3v3 & MAMuJoCo  & Marathon\\ \midrule

Heterogenous Support  
& \ding{51}   & \ding{51}   & \ding{51}     & \textbf{--} & \ding{51}   & \ding{51}   & \ding{51}     \\
Large-Scale Support
& \textbf{--} & \textbf{--} & \textbf{--}   & \textbf{--} & \textbf{--} & \ding{51}   & \textbf{--}   \\
Multi-Team Support          
& \textbf{--} & \textbf{--} & \textbf{--}   & \textbf{--} & \textbf{--} & \textbf{--} & \textbf{--}   \\
Mixed-Game Support 
& \textbf{--} & \ding{51}   & \ding{51}     & \textbf{--} & \textbf{--} & \textbf{--} & \textbf{--}   \\

3D Physics Engine             
& \ding{51}   & \ding{51}   & \ding{51}   & \ding{51}   & \ding{51}   & \ding{51}     & \ding{51}\\

Fully Open Source 
& \ding{51} & \textbf{--} & \textbf{--} & \ding{51}   & \textbf{--} & \ding{51}     & \textbf{--} \\

All Elements Customizable 
& \textbf{--} & \textbf{--} & \textbf{--} & \textbf{--} & \textbf{--} & \textbf{--}   & \textbf{--}  \\

Controllable Time-flow Speed 
& \textbf{--} & \textbf{--} & \textbf{--} & \textbf{--} & \textbf{--} & \textbf{--} & \textbf{--}\\ 

Rendering Training         
& \textbf{--} & \textbf{--} & \textbf{--} & \textbf{--} & \textbf{--} & \textbf{--} & \ding{51}\\

\bottomrule
\end{tabular}
} 
\end{table}


The domain-specific simulation environments for MARL can be further broadly categorized into two types: those with physics engines and those without. Here, physics engines refer to a suite of tools capable of simulating the physical laws inherent in real-world tasks~\citep{templet2021game}. Given that game engines also aim at reincarnating the real-world elements into the digital world~\citep{vohera2021game}, environments leveraging game engines are classified under the physics engine category. 

Among the environments without physics engines, MPE~\citep{MPE} utilizes a simple rule-based particle world to simulate multi-agent tasks such as predator-prey and cooperative navigation. MAgent~\citep{MAgent}, grounded in a grid world, facilitates simulations involving the aggregation and combat of pixel-block agents, notable for its ability to support large-scale multi-agent settings. The two environments mentioned above are based on the state transition laws of particle worlds and particle interactions. Although they are completely open-source and their task elements are relatively easy to modify, their scenarios are overly simplistic and lack realism.

Hanabi~\citep{Hanabi} provides a multiplayer card game scenario, which is commonly used in MARL research based on opponent modeling. However, the overly narrow theme prevents it from further simulating tasks involving heterogeneity, large scale, and mixed strategies. Neural MMO~\citep{NerualMMO} is developed in a 3D grid world derived from massively multiplayer online games, supporting large-scale multi-agent simulations over long time horizons.

Gobigger~\citep{GoBigger}, based on a ball world concept, stands out for enabling simulations involving collaboration and competition among multiple teams. 
However, similar to MPE and MAgent, their particle-based 2D environments fall significantly short of simulating the real-world complexities of 3D scens.

As for the environments with physics engines, GRF~\citep{GRF} is built upon the GameplayFootball simulator~\citep{GameplayFootball}, creating a highly realistic football match setting that allows agents to simulate the behaviors of human players. However, it does not support large-scale scenarios, multi-team training, and mixed multi-agent gameplay. Moreover, although its environment interface and underlying engine are open-source, the underlying engine is not suitable.

SMAC~\citep{SMAC} and SMACv2~\citep{SMACv2} are developed based on the popular video game StarCraft II, constructing a multi-agent micromanagement environment where each agent controls individual units to complete adversarial tasks. Despite their widespread use, the fact that their underlying games and engines are not fully open-source limits further expansion, confining their built-in tasks to battle-type game scenarios only.

Hide-and-Seek~\citep{Hide-and-Seek} has set up a series of multi-agent curriculum learning scenarios, such as hide and seek, based on a 3D engine. However, its theme is too singular, making it impossible to simulate tasks involving heterogeneity, large scale, multiple teams, etc., and it does not allow for customization of all task elements.
Hok3v3~\citep{HoK3v3}, specifically designed for heterogeneous multi-agent tasks, is based on the Honor of Kings engine, with agent action spaces consistent with those of human players engaging in the real game. However, it only supports heterogeneous multi-agent scenarios (3VS3) and does not have an open-source underlying game and related engine.

MAMuJoCo~\citep{MAMuJoCo} is developed using the Mujoco physics engine~\citep{Mujoco}, where multiple agents each control different joints to collaboratively manage the movements of a single robot. However, all of the multi-agent scenarios are fully cooperative and do not support large-scale multi-agent tasks.

Marathon Environment~\citep{Marathon} is developed using the Unity3D engine, supporting multiple agents learning complex movements such as running and backflipping. The built-in tasks are relatively simple and are unable to simulate large-scale, multi-team, and mixed multi-agent gameplay tasks. Moreover, its underlying engine, Unity3D, is not fully open-source, thus preventing comprehensive modifications from the bottom to the top layer.

JaxMARL~\citep{JaxMARL} integrates numerous MARL environments together and has re-implemented these environments using JAX technology, enabling them to support efficient, GPU-based parallel computing. However, to support pure GPU parallelism, some environments in JAXMARL have lost their original CPU-based underlying physical engines. Moreover, as a collection of environments that integrates multiple basic environments, it does not support multi-team multi-algorithm training, nor does it support controllable time-flow simulation and cross-platform real-time rendering.

It is evident that environments without physics engines are adept at simulating challenging tasks designed to push the limits of existing algorithms. In contrast, environments equipped with physics engines offer greater potential for real-world applications but are constrained in terms of academic flexibility.
Our goal is to develop an environment that not only has practical application potential but also fully leverages scalability, ultimately leading to the creation of Unreal-MAP.

\section{Unreal-MAP Details}
\label{App:Unreal-MAP Details}
\subsection{Architecture of Unreal-MAP}

Unreal-MAP utilizes a hierarchical design that consists of five layers, all of which are open source. As shown in Figure~\ref{fig:Framework of Unreal-MAP}, the first layer of Unreal-MAP is the native part of the Unreal Engine, including the physics engine, rendering engine, AI engine, and a range of 3D assets. We build the entire Unreal-MAP based on the open-source version of UE, making modifications to some of the native modules. For instance, the original AI detection system for agents in UE was very inefficient in large-scale scenes. Unreal-MAP optimizes the detection of multiple entities by incorporating tensor operations and eliminating redundant checks.

The second layer of Unreal-MAP comprises the underlying systems and programming specification, all implemented in C++. The time control system and task system in this layer ensure the correct initiation and termination of simulation episodes, guaranteeing the precision of simulation time steps and the reproducibility of experimental results. Other components of this layer define the specification for all base classes, communication, and debugging within Unreal-MAP.

The third layer consists of three fundamental classes implemented using Blueprints. The agent class defines all entities that can be controlled by algorithms, while the entity actor corresponds to all environmental entities that do not make decisions. Classes derived from these two form all the entity elements within a task scenario. The abstract class acts as a bridge, connecting the underlying systems to the highest Python-based layer, facilitating communication, debugging, action updates, and observation feedback.


The fourth layer of Unreal-MAP consists of advanced functional modules, implemented using Blueprints. These modules allow for the modification of various attributes of agents, including appearance, perception, action sets, movement, and navigation, enabling the development of diverse types of agents. Moreover, by leveraging the abundant resources in the Unreal community, the map construction module facilitates the creation of new maps and even the importation of real-world maps. The entity construction module aids in developing complex environmental entities, such as altering the kinematic model of missiles launched by drones.

\begin{figure}[!h] 
    \centering
    \includegraphics[width=0.85\linewidth]{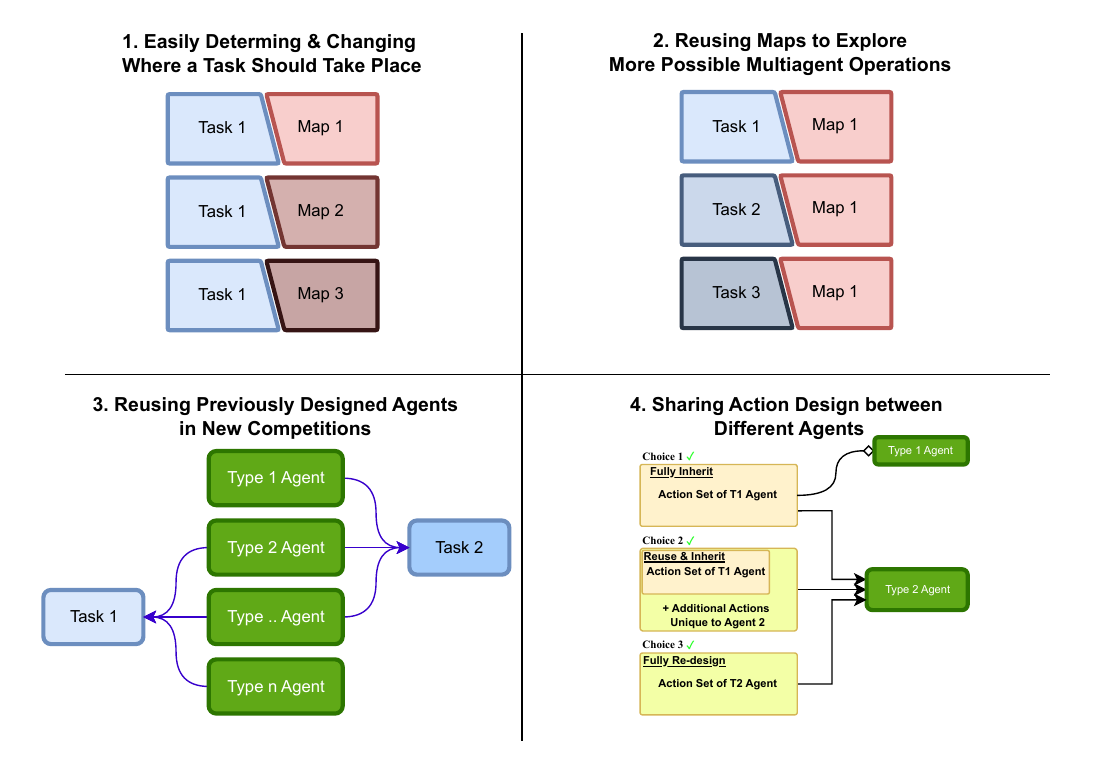}
    \caption{One of the advantages of Unreal-MAP framework is the isolation of maps, task and agents, making it possible to reusing existing modules to develop new environment for Reinforcement Learning studies.}
    \label{fig:advantage_1}
\end{figure}

The fifth layer serves as the interface for interaction between Unreal-MAP and algorithms, all implemented in Python. This interface adheres to the gym~\citep{Gym} specification, encompassing basic functions like reset, step, and done, and supports the customization of agent-level observation and reward functions. Attributes such as agent size, initial position, detection range, and health are directly encapsulated within the agent initialization function, allowing for easy modification. As shown in Figure~\ref{fig:advantage_1}, the selection of agents, tasks, and maps in Unreal-MAP are independent. Users can customize the types, numbers, and teams of agents in a task and switch maps flexibly.

From the perspective of designing and utilizing a MARL simulation environment, users only need to focus on the fourth and fifth layers of Unreal-MAP. In most cases, users can directly customize MARL tasks by modifying the built-in scenarios and agent parameters through the fifth layer. If there is a need to develop new scenarios or further develop existing ones, users can also easily develop through the graphical programming approach provided in the fourth layer. Such hierarchical design of
Unreal-MAP significantly reduces the burden of customizing tasks.

\subsection{Time in Unreal-MAP}
\label{Time}
Time is one of the most important factor in simulations.
There are two different type of time in Unreal-MAP:
\begin{enumerate}
    \item Real World Time $t_{{real}}$. The actual time of our world.
    \item Simulation Time $t_{sim}$. The time in the simulated virtual world.
\end{enumerate}

It is inevitable that simulation speed (from the perspective of $t_{real}$) will be influenced by factors such as CPU frequency, GPU performance, policy neural network size, machine workload, etc.
As a result, Unreal-MAP decouples simulation time flow 
therefore has achieved flexible control of simulation time 
\begin{enumerate}
    \item Unreal-MAP allows researchers to slow down simulation time by setting a time dilation factor, extending a second in the simulation multiple times to render details of agents in slow motion.
    \item Unreal-MAP allows researchers to accelerate simulation time by setting the same time dilation factor (before reaching the hardware limitation). Gathering large amount of samples is necessary in most RL tasks. Accelerating computation is the primary ways to achieve this goal.
\end{enumerate}
Unreal-MAP guarantees that the simulation results will not be influenced by time dilation factor, hardware or workload.
For instance, as long as the random seed remains identical, same agent trajectories are expected:
1) regardless of whether we choose to enable GPU to accelerate neural network computation.
2) regardless of whether we choose to simulate agents slowly or rapidly by setting different time dilation factors.


There are three global time-related settings to adjust in Unreal-MAP.

\textbf{Decision time interval.} From the perspective of agents in the simulated environment, agents will have a chance to act once every $t_{\text{sim}}^{\text{step}}$. 
Alternatively, $t_{\text{sim}}^{\text{step}}$ is also the time interval between each RL step.
$t_{\text{sim}}^{\text{step}}$ is usually a short period with a default value 0.5s.
Nevertheless, for tasks such as flights that last hours in a episode, $t_{\text{sim}}^{\text{step}}$ should be increased accordingly.

$t_{\text{sim}}^{\text{step}}$ does NOT has directly relationship with how long a RL step will actually take in the real world. More specifically, a team can take as long as necessary to compute the next-step action after receiving observation, meanwhile the simulation time flow freezes until all teams have committed agent actions. In extreme situations, algorithms can spend hours to update large policy networks and the simulated agents will not be influenced by this delay.

\textbf{Baseline Frame Rate.} Baseline Frame Rate $t_{\text{sim}}^{\text{fr}}$ determines how many frames to compute for each simulation second in Unreal-MAP.
As an example, when $t_{\text{sim}}^{\text{fr}}=30$, the simulation will proceed (tick) $\frac{1}{30}$s after each frame. Important computation such as  collision detection and agent dynamic update are performed in each of these frames.
As an example, let $t_{\text{sim}}^{\text{step}} = 0.5$ and $t_{\text{sim}}^{\text{fr}}=30$, under this circumstance $15$ ticks will be performed between each RL step.
Similarly, $t_{\text{sim}}^{\text{fr}}$ does NOT have direct relationship with the real world time flow.

\textbf{Time Dilation Factor.} In Unreal-MAP, Time Dilation Factor $t_{\text{real}}^{\text{df}}$ is the sole bridge between simulation time flow and real world time flow. In reinforcement learning, there are three typical cases that involves the control of time in simulation:

\begin{enumerate}
    \item Task Development and Evaluation. In this case, it is demanded that simulation time flows at a normal speed to observe the interaction of agents. A dilation factor $t_{\text{real}}^{\text{df}} \approx 1$ will synchronize simulation time flow with the real world time flow.
    \item Slow Motion. In this case, it is required that the simulation runs slowly to allow human observers to diagnose issues in multi-agent cooperation. Changing the dilation factor $t_{\text{real}}^{\text{df}} < 1$ will slow down the simulated world accordingly.
    \item Training. In this case, it is demanded that simulation runs as fast as possible to collect training data. Unreal-MAP will attempt to accelerate the simulation until reaching the $t_{\text{real}}^{\text{df}}$ threshold. If not possible due to hardware, the simulation will still proceed at the fastest possible simulation speed.
\end{enumerate}

\subsection{Underlying Optimization in Unreal-MAP}
\label{Underlying Optimization}
This section focus on the specific limitations associated with Unreal-MAP and how we have tackled these challenges, when utilizing the Unreal Engine for handling large-scale multi-agent tasks.

The computation within each agent is automatically managed in parallel by the Unreal Engine, thus this aspect does not contribute to the limitations of large-scale simulations. During our testing phase, we have identified three primary bottlenecks:
\begin{enumerate}
    \item \textbf{Agent-wise Perception Modules:} The challenge arises when managing a large number of agents (e.g., 200 agents), each with potentially different perception distances and shapes (e.g., cone, sphere, etc.). This necessitates the computation of a 200x200 perception matrix. To address this issue, we have implemented several strategies, including:
    \begin{itemize}
        \item Batched distance computation utilizing the xtensor library.
        \item Parallel processing of perception computations.
        \item Redesign of the existing Unreal Engine built-in agent perception pipeline to optimize performance.
    \end{itemize}
    \item \textbf{Memory Leak:} When handling many agent and non-agent objects, memory leaks can be observed. These leaks can degrade system performance over time, especially in large-scale simulations where numerous objects are continuously created and destroyed. To address this issue, we have enhanced the recycling mechanism for agent and non-agent objects within episodes to ensure no memory leaks.
    \item \textbf{Inter-Process Communication (IPC) Time Consumption:} IPC between Python and the UE simulation becomes a bottleneck as the scale of the simulation increases. While IPC costs can be ignored for smaller team sizes due to efficient handling by the operating system, they become significant when the number of agents exceeds 100, leading to a substantial increase in IPC package size. Our strategies for mitigating this issue include:
    \begin{itemize}
        \item Utilization of fast compression algorithms, such as lz4, to reduce data size.
        \item Development of shared memory implementations to further enhance efficiency and reduce IPC time consumption.
    \end{itemize}
\end{enumerate}

\subsection{Research Workflow of Unreal-MAP}
\label{WorkFlow}

This section focuses on answering how much effort should a new developer invest in integrating their own environment/problem into Unreal-MAP.
To answer this question, we first clarify the basic workflow for a new developer using Unreal-MAP, which includes: \textit{Environment Setup}, \textit{Task Development}, \textit{Algorithm Training}, \textit{Result Rendering}, and \textit{Feedback for Further Development}.

Notably, the effort spent on \textit{Result Rendering} is independent of the specific tasks. Developers can master the full set of Unreal-MAP rendering mechanisms by spending less than an hour watching our tutorial videos, as no secondary development on the rendering mechanism is required, only straightforward operations. For \textit{Environment Setup}, \textit{Task Development}, and \textit{Algorithm Training}, we will analyze both the simplest and the most complex cases.

In the simplest case, developers do not need to make any changes on the UE side. For \textit{Environment Setup}, they only need to deploy our pre-configured Docker image, which includes compiled binary files, HMAP code, and a series of environment configurations, all achievable within minutes. For \textit{Task Development}, utilizing built-in scenarios and agents through the Python interface layer allows for quick customization of tasks. As for \textit{Algorithm Training}, users can choose their familiar third-party code frameworks and directly copy the algorithm code into HMAP's third-party code folder.

In the most complex case, such as when users need to develop entirely new scenarios and even conduct physical experiments for validation, the \textit{Environment Setup} step requires installing Unreal Editor and our modified Unreal Engine (the fundamental layers of Unreal-MAP), which can typically be completed within two hours under tutorial guidance. For \textit{Task Development}, users can design and develop agents and maps using the rich resources available in the Unreal Game Store\footnote{\url{https://www.fab.com}.}, usually within a week. If further integration with physical experiments is needed, the development of kinematics and dynamics for agents, action design, and behavior tree design may require more effort. However, thanks to UE's simple and practical Blueprint language, such tasks can typically be completed within a month. \textit{Algorithm Training} in complex cases is similar to the simple case, where users only need to merge the code from their familiar third-party frameworks into HMAP.

\section{HMAP Details}
\label{app:HMAP Details}

\begin{figure}[!h] 
    \centering
    \includegraphics[width=0.99\linewidth]{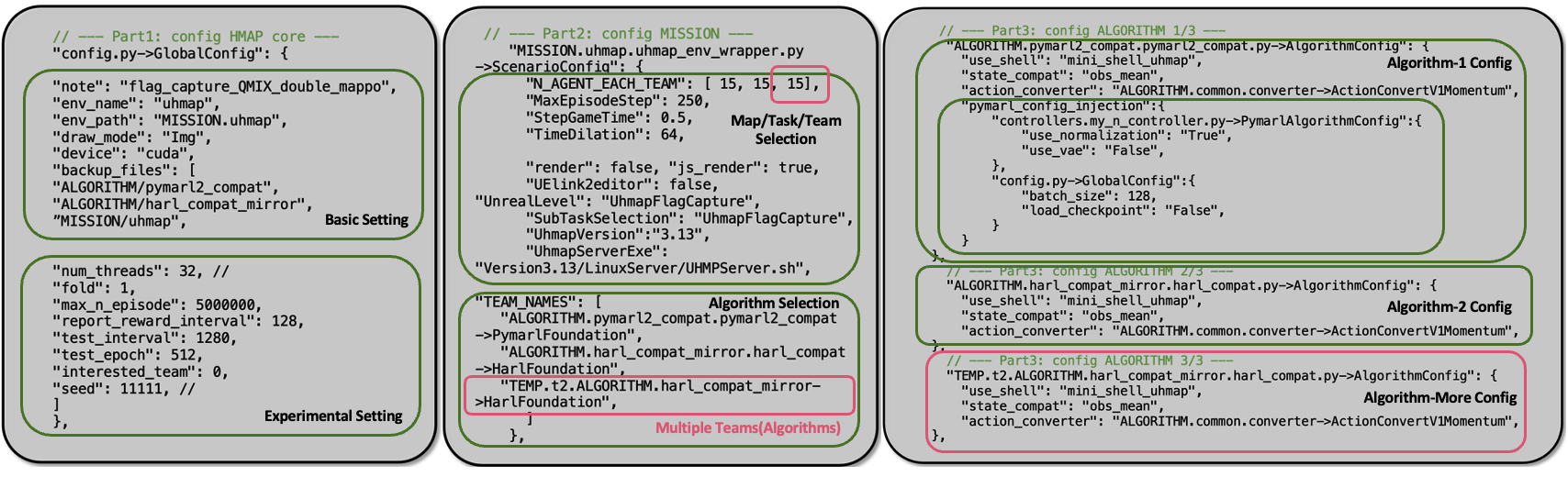}
    \caption{Parameter configuration of HMAP. This is an example under the \textit{flag\_capture\_double\_mappo} task using the QMIX algorithm. To facilitate multi-team training, users only need to add teams and designate their respective algorithms in the mission config, and append the corresponding algorithmic parameters in the Algorithm config.
}
    \label{fig:HMAP}
\end{figure}

The utilization of HMAP is straightforward, necessitating only a Docker container, a configuration of parameters, and execution of a single command to deploy a specified algorithm into a designated scenario. The parameter configuration files of HMAP exemplify its modular design, including three categories: core config, mission config, and algorithm config. Figure~\ref{fig:HMAP} illustrates a configuration file for the QMIX algorithm under the \textit{flag\_capture\_double\_mappo} task. \textbf{Core config} comprises basic settings and experimental settings, where the former allows for the specification of file and path for experiment storage, and the latter includes parameters relevant to the experiment such as the number of parallel task environments, testing intervals, and random seeds. \textbf{Mission config} includes selections for the simulation environment and deployed algorithm. Upon selecting Unreal-MAP as the simulation environment, users can make further selections regarding maps, tasks, and teams, as well as choose between training, rendering, or a mode that combines both training and real-time rendering. \textbf{Algorithm config} is composed of the algorithmic parameters set for each team.

In the multi-team training of HMAP, the observations, actions, and reward data for each algorithm are processed separately and efficiently, ensuring that the policy executing and training for each team are independent. HMAP accommodates sequential and parallel updates of multiple team policy according to hardware performance variations, with the update sequence having no adverse impact on the effectiveness of algorithm training.

With HMAP, users can conveniently specify the number of teams and freely assign algorithms to each team. For instance, Figure~\ref{fig:HMAP} demonstrates the setting of three teams, where Team-1 is assigned QMIX from the PyMARL2 framework, and Team-2 and Team-3 are designated MAPPO from the HARL framework. HMAP allows multiple teams to utilize the same algorithm module without affecting the normal construction of buffers and network updates. It is achieved by adding a prefix keyword like ``TEMP.t2" to the additional configurations of the same algorithm. Theoretically, as long as computational resources are sufficient, Unreal-MAP and HMAP can support an arbitrary number of teams, each allocated with different algorithms in a same scenario, with the updates of different algorithms not interfering with each other.

\section{Scenario Details}
\label{app:Scenario Details}

\textbf{Metal Clash.}
This scenario is designed for heterogeneous multi-agent tasks and large-scale multi-agent tasks. It involves an SMAC-style competition between two teams of agents. Each team can be controlled by either rule-based or learning-based algorithms. The objective of the ally team is to eliminate as many enemy agents as possible while preserving more ally agents.

Metal Clash offers three types of basic agents: missile cars (for ground and air attacks), laser cars (for ground attacks), and support drones (for attacks and supports). Missile cars can attack ground or aerial units with missiles and have a long range, but they move slowly. Laser cars excel at close-range combat, using lasers to damage ground units. Support drones, as aerial units, have a faster movement speed and can restore the health points of allied missile cars and laser cars. They can also attack opponents with smaller firepower but have lower HP.

The physical attributes of each basic agent, such as movement speed, size, HP, observation radius, etc., are exposed as configurable parameters at the task construction interface. Users can conveniently modify all the parameters, thereby creating a variety of heterogeneous agent types that far exceed the original three. Agents can sense their neighborhood allies and enemies within the perception range, and the information of perceived agents is concatenated in the observation space. Due to the height advantage, the flying agents have a much larger perception range than ground agents. Agents have rich options of actions. Besides the idle and moving actions, agents can choose a patrol-moving action to search enemies, select visible opponents to attack, or toggle their micro-management strategy (such as whether agents are allowed to peruse opponents after receiving future attack action). Furthermore, users can freely designate the types and corresponding numbers of agents in both allied and enemy teams, thus controlling the nature and difficulty of the task.

\textbullet \quad \textit{Observation}

Three types of agents are intentionally designed differently to reflect the heterogeneity of this scenario. We define the distance unit of the unreal engine as \textit{u}. Missile car have a maximum movement speed of 500u per second, an attack power of 1, an attack range of 1000u, and 150 HP. Laser cars have a maximum movement speed of 800u per second, an attack power of 1, an attack range of 500u, and 100 HP. Support drones have a maximum movement speed of 1000u per second, an attack power of 1/6, an attack range of 1700u, and 50 HP. The observation capabilities of three agents are shown in Table ~\ref{Tab: Observation of Metal Clash}. The observation structure refers to the composition of what an agent observes, where the number 1 represents its own observation, and the subsequent two numbers indicate the maximum number of allied agents and enemy agents that can be observed. For example, missile cars have an observation range of 2500u, and their observation structure is [1,8,8]. This indicates that missile cars can observe information about up to 10 
ally agents and 10 foe agents within a range of 2500u. Additionally, the information for each observed agent is a 20-dimensional vector, with vectors for invalid entities filled with zeros. Therefore, the observation dimension for the missile vehicle is (17)*20.

\begin{table}[ht]
\centering
\caption{Observation Capabilities of three base agents in Metal Clash.}
\label{Tab: Observation of Metal Clash}
{\small 
\renewcommand{\arraystretch}{1.0} 
\begin{tabular}{cccc}
\toprule
Agent & Observation Range & Observation Structure & Observation Dimension\\ \midrule
Missile Car & 2500u & [1,8,8] & (17)*20\\ 
Laser Car & 2000u & [1,5,5] & (11)*20 \\
Support Drone & 2500u & [1,10,10] & (21)*23\\
\bottomrule
\end{tabular}
} 
\end{table}

\textbullet \quad \textit{Action}

All three types of agents have nine common actions: moving in four directions, staying still, targeting foe agents within the defense circle, fleeing, etc. On this basis, each type of agent can perform special actions. For example, support drones can choose to restore the health of ally agents within their support range or choose to attack foe agents. Missile cars can choose to attack all units, while laser cars can only choose to attack ground units.

\textbullet \quad \textit{Reward}

Regarding reward settings, when an agent from our team or the enemy team is destroyed, the entire team receives a penalty of 0.05 and a reward of 0.1. At the end of an episode, the team with the higher total remaining HP wins, receiving a reward of 1.0, while the losing team receives a penalty of -1.0. In the event of a tie, both teams receive a penalty of -1.0.

\textbf{Monster Crisis.}
This scenario consists of a monster and several mushroom agents. Users can adjust the task's difficulty by altering the monster's defensive capabilities and the number of agents involved. In the most challenging cases, agents must form a precise formation beyond the monster's defense range and launch a swift, simultaneous attack to just manage to kill the monster.

Regarding reward settings, the entire team receives a positive reward only if the agents successfully kill the monster; there are no rewards or penalties in other cases. Similar to the \textit{Metal Clash} scenario, both the tower and the agents have spherical perceptual space centered around themselves. Agents can choose from idle, move to a certain direction, or attack in their action space.

\textbullet \quad \textit{Observation}

In \textit{Monster Crisis}, each agent has the same observation space. They can observe information such as the ID, HP, position, and maximum speed of themselves and the position of the monster.

\textbullet \quad \textit{Action}

In \textit{Monster Crisis}, agents have six possible actions to choose: moving in four directions, maintaining the current action, and staying still. Additionally, each agent has an action to collide with the monster.

\textbullet \quad \textit{Reward}

This scenario is set up for sparse team rewards. Rewards are given only when all agents cooperate to kill the monster. Specifically, agents can receive a reward of value 1 only when they defeat the monster.



\textbf{Flag Capture.}
This scenario allows for competition among more than two teams. At the end of an episode, only the team that holds the flag for the longest duration wins. Since each team begins with an equal number of agents, the first team to capture the flag does not guarantee victory, as teams must carefully consider the balance of power and strategic play among multiple teams to receive the most rewards. In this scenario, agents are not equipped with weapons and cannot eliminate other agents. Consequently, agents do not have attack actions. Moreover, the agent's perceptual space is conical rather than spherical.

\textbullet \quad \textit{Observation}

In \textit{Flag Capture}, each agent has the same observation space. They can observe information such as the ID, HP, position, and maximum speed of ally or foe agents within the observation range.

\textbullet \quad \textit{Action}

Each agent has a constant speed. In the two-dimensional plane, there are eight discrete actions to choose from, each representing a direction spaced 45 degrees apart. When the team is close enough to the flag, the agent nearest to the flag will pick it up. To prevent other teams from approaching and capturing the flag, it is necessary to target the agent that are near the flag.

\textbullet \quad \textit{Reward}

When a flag is picked up by an agent, the team to which the agent belongs receives a reward of 0.005. At the end of the episode, the team that has held the flag for the longest time will receive a reward of 1.0.

\textbf{Navigation Game.}
The scene includes various obstacles and walls, along with two landmarks.  If the air navigator stays over any landmark for a certain period (10 seconds), the navigator team is deemed to have won.  

\textbullet \quad \textit{Observation}

In this scenario, agents can perceive the location and status of the target area regardless of the distance, yet can only sense opponents within agents' perception range.

\textbullet \quad \textit{Action}

All three types of agents have nine common actions: moving in four directions, staying still, fleeing, etc. 

\textbullet \quad \textit{Reward}

The rewards in this scenario include dense rewards and sparse rewards. For the navigator team, the dense reward is the average distance of air navigators to the nearest landmark divided by 10,000, which serves as a guiding reward. The sparse reward occurs at the end of the episode, with a reward of 1.0 for winning and a penalty of 1.0 for losing. The rewards for the keeper team are the opposite of those for the navigator team.






\section{Unreal-MAP's Efficiency and Computational Resource Consumption}
\label{app:Eff}

As a simulation environment based on a 3D physical engine, Unreal-MAP boasts high simulation efficiency. It is well-designed to adapt to and fully utilize various types of computing resources. Unreal-MAP can be deployed on computing systems entirely devoid of GPUs for algorithm training and supports the full utilization of uneven computing resources. It can operate in a single-threaded manner as well as support multiple parallel environments. Additionally, with the feature of time dilation factors, Unreal-MAP can not only improve simulation efficiency by increasing the number of parallel processes but also control the simulation speed of each process to make full use of computing resources (using more CPU utilization under the same memory and GPU memory), a functionality not available in other simulation platforms.

In this section, to verify Unreal-MAP's efficiency and adaptability to various computing resources, we conducted a series of experiments on Unreal-MAP's efficiency index and resource consumption indices. The efficiency index adopted was TPS, i.e., the number of virtual timesteps run in a real second; the resource consumption indices included CPU utilization, memory occupancy, and GPU memory occupancy. All experiments were conducted on a Linux server equipped with an AMD7742 CPU (maximum frequency 2.25GHz) and NVIDIA RTX3090 GPUs. To ensure fairness, all experiments tested the QMIX algorithm on the \textit{metal\_clash\_5sd\_5mc} task. The data points for all indices were obtained by averaging the results of five experiments. At the beginning of each experiment, the server was maintained in an idle state, executing only the essential system processes.

From Figure~\ref{fig:Eff}, the following conclusions can be drawn:
\begin{enumerate}
    \item With a constant number of parallel environments, TPS and CPU utilization are roughly proportional to the time dilation factor, but this proportional relationship degrades into a positive correlation when the time dilation factor reaches a certain threshold (limited by the CPU's clock speed).
    \item With a constant number of parallel environments, changing the time dilation factor almost does not affect memory occupancy and GPU memory occupancy.
    \item With a constant time dilation factor, CPU utilization and TPS are roughly linearly related to the number of parallel environments; memory and GPU memory occupancy are positively correlated with the number of parallel environments.
\end{enumerate}

The above conclusions mean that under limited memory resources, training efficiency can be improved by increasing the time dilation factor to fully utilize CPU resources; similarly, under limited CPU computing resources, reducing the time dilation factor and increasing the number of processes can avoid the waste of computing resources.

In fact, when the number of processes is 8 and the time dilation factor is 32, training 1024 episodes on the \textit{metal\_clash\_5sd\_5mc} task takes less than 2 minutes. This means that under such parameter settings, this server can simultaneously support 50 such tasks (each with 20 agents) and complete all training tasks (100k episodes) within 3 hours. In special cases, the number of parallel processes can be further increased to improve training efficiency. When the number of processes reaches 128 and the time dilation factor is set to 32, the TPS can reach 1000+, and the training task can be completed in about an hour.

It is important to emphasize that TPS here counts the number of virtual Unreal-MAP timesteps per real second. Considering this is a simulation of 20 agents, and each timestep in Unreal-MAP undergoes 1280 frames of calculations for environmental dynamics and kinematics to maintain fine state transitions (details in Appendix C.2), this is already highly efficient computation.

\section{Checklist of Ethics}
\label{Ethics}
We have finished a checklist to facilitate discussions on the ethical considerations of artificial intelligence involved in our work. This checklist~\citep{ThornyRoses} addresses the potential impacts of various artefacts in the field of artificial intelligence.

\textbf{C1 Did you explicitly outline the intended use of scientific artefacts you create?}

Yes. The scientific artefacts we have created are Unreal-MAP and HMAP. The former is a general platform developed based on the Unreal Engine, designed with a layered architecture to enable users to conveniently develop various realistic 3D multi-agent simulation environments. The latter is an experimental framework that is highly compatible with Unreal-MAP, characterized by its support for multi-team multi-algorithm training, and compatibility with existing classic simulation environments and algorithms from third-party frameworks. The purpose of developing Unreal-MAP and HMAP is to enable users to develop simulation environments that meet their needs (including sim-to-real transfer and new research ideas) in the field of MARL, and to rapidly deploy algorithms to validate ideas, thereby promoting the development of the MARL field.

\textbf{C2 Can any scientific artefacts you create be used for surveillance by companies or governmental institutions?}\\

No. As a simulation environment and experimental framework in the MARL field, Unreal-MAP and HMAP are unrelated to surveillance by companies or governmental institutions.

\textbf{C3 Can any scientific artefacts you create be used for military application?}\\

The motivation and details of creating Unreal-MAP and HMAP are unrelated to any military applications. However, it must be emphasized that although the design inspirations, virtual materials, and physical materials used in this work are unrelated to military applications, there is a risk if our work is applied to MARL policy training for military purposes. Therefore, on one hand, we call on the open-source community to strengthen the regulation of military application materials and urge users to refrain from using Unreal-MAP for military purposes. On the other hand, we also plan to set up keyword detection within the UE side, so that users with impure motives designing military application-related scenarios will not be able to use the functions of Unreal-MAP.

\textbf{C4 Can any scientific artefacts you create be used to harm or oppress any and particularly marginalised groups of society?}\\

The motivation and details of creating Unreal-MAP and HMAP are unrelated to harming or oppressing any particularly marginalized groups of society. In fact, our environment and experimental framework are suitable for users under various computing resources, and are compatible with various system platforms.

\textbf{C5 Can any scientific artefacts you create be used to intentionally manipulate, such as spread disinformation or polarise people?}\\

The motivation and details of creating Unreal-MAP and HMAP are unrelated to intentional manipulation, such as spreading disinformation or polarising people. However, it must be emphasized that although the experiments and demonstrations based on Unreal-MAP are unrelated to this. If the simulation environments developed using Unreal-MAP can be used to generate realistic false scenarios, there is a risk of being maliciously used to create and spread false information. Therefore, we call on the open-source community to participate in regulation, establish a reporting mechanism, and we will add educational materials for users in the usage tutorials, emphasizing the ethical responsibility of using simulation environments and raising users' ethical awareness.

\textbf{C6 Did you access your institution’s or other available resources to ensure limiting the misuse of your research?}\\

Yes, we have accessed our institution to ensure limiting the misuse of our research, including but not limited to the promotion, use, and modification of this work.

\textbf{C7 have you been provided by your institution with ethics training that covered potential mis-use of your research?}\\

Yes, we are confident that our institution has provided sufficient ethics training.

\textbf{C8 Were the scientific artefacts you created reviewed for dual use and approved by your institution’s ethics board?}\\

Yes, the scientific artefacts we created have been reviewed for dual use and approved by our institution's ethics board.

\section{Physical Experiment Details}
\label{sec:Physical Experiment Details}

\begin{figure}[!h] 
    \centering
    \includegraphics[width=0.99\linewidth]{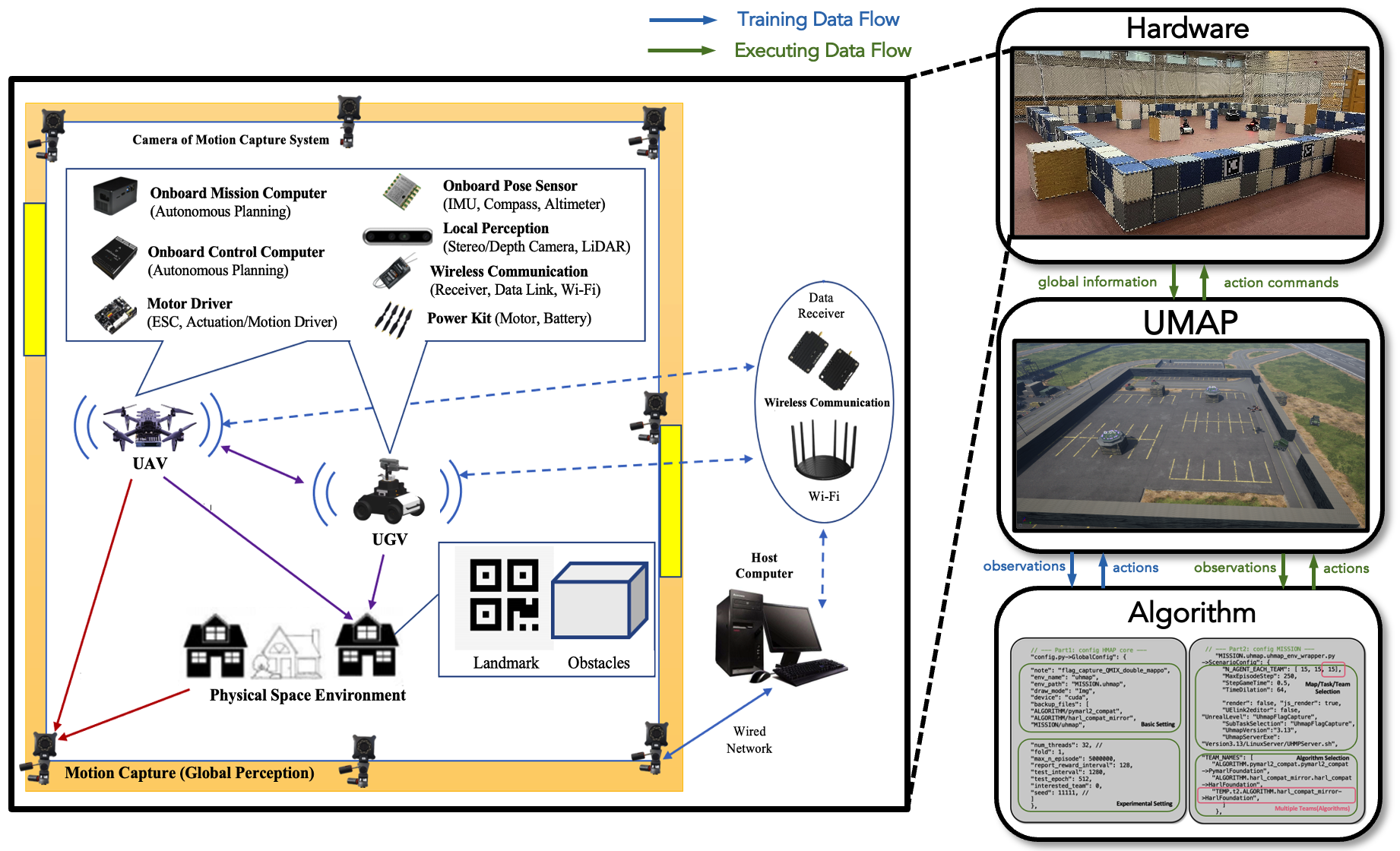}
    \caption{The \textbf{Algorithm}-\textbf{Unreal-MAP}-\textbf{Hardware} framework.
}
    \label{fig:AlgorithmUnreal-MAPHardware}
\end{figure}

As shown in Figure~\ref{fig:AlgorithmUnreal-MAPHardware}, the overall framework of the physical experiment includes three components: the algorithm side represented by HMAP, the virtual environment side represented by Unreal-MAP, and the hardware-based real environment side. During the training phase, HMAP and Unreal-MAP communicate through the TCP protocol, exchanging observations and action information of the environment, completing the training tasks on the same host server/computer. During the executing phase, Unreal-MAP needs to maintain communication with not only  HMAP but also with the communication system in the real environment side through the TCP protocol, transmitting global observation information and decoded action information. In addition to the communication system, the real environment side also includes an action capture system, several UAVs and UGVs, landmarks and obstacles, and a host computer. The motion capture system transmits global information (the position, speed of all entities) to the host computer through a wired network, which receives local observation information (such as the first-person view from cameras) from UGVs and UAVs through a wireless communication module and sends commands to them.

The UGVs and UAVs in the real environment have autonomous planning and control capabilities. They can receive the information of target position or target speed from the communication module and complete commands through two-dimensional and three-dimensional PID control. Unreal-MAP also replicates their PID kinematics. UGVs are also equipped with cameras, which send the first-person view information to Unreal-MAP. Unreal-MAP simulates their viewpoints, combined with the global information from the motion capture system, to create simulated observation information under partially observable conditions for HMAP. In the simulated environment training, the simulated UAVs also have limited viewpoints, being able to observe entity information only within a specified range and angle.

\section{Experimental Details}

\subsection{Hyperparameter Details}
\label{sec:Hyperparameter Details}

In this part, the common hyperparameters used for algorithms and tasks are described. We present the hyperparameters used for actor-critic-based algorithms in Table~\ref{Tab: hyperparameters1} and for value-based algorithms in Table ~\ref{Tab: hyperparameters2} across all tasks. Other unspecified hyperparameters of algorithms remain at their default settings. The hyperparameters used for tasks are shown in Table ~\ref{Tab: hyperparameters3}.

\subsection{Additional Experiments and Analysis}
\label{app:Additional Experiments and Analysis}
\textbf{Heterogeneous Tasks}. 
The result plotted in Figure~\ref{fig:15Result} reveals several trends. Apart from QPLEX, actor-critic-based algorithms generally outperform value-based algorithms. In actor-critic-based algorithms, MAPPO performs better, even being the best algorithm in the most difficult task, and HAPPO is weaker than MAPPO across all four tasks, which is different from previous research. In value-based algorithms, QPLEX is the best, which outperforms all actor-critic-based algorithms in \textit{metal\_clash\_5lc\_5mc}. However, it is discovered that the effectiveness of QPLEX significantly declines as the level of heterogeneity in the task increases. Furthermore, experiments without parameter sharing are conducted, and it has been found that actor-critic-based algorithms with parameter sharing outperform those without parameter sharing. Since the agent ID is already included in the observations, this enables differentiation among the trained policies.

\textbf{Large-Scale Tasks}. 
Similar to heterogeneous tasks, actor-critic-based algorithms still outperform value-based algorithms. MAPPO is the most outstanding algorithm due to its superior capability for parameter sharing, which is primarily reflected in its faster and more stable training performance. This advantage is particularly evident in \textit{metal\_clash\_hom\_100} and the highly heterogeneous \textit{metal\_clash\_het\_100}, where MAPPO demonstrates a significant lead. For value-based algorithms, the performance of QPLEX is the best, but it also deteriorates rapidly with the increase in scale and heterogeneity. Furthermore, the training of HAPPO is very unstable, which may be related to its updating of policies in a random order. In tasks with 100 agents in the team, HATRPO freezes up and fails to produce results, because the computational burden of HATRPO is so large that it exceeds the computing capacity of the server. Apart from MAPPO and QPLEX in \textit{metal\_clash\_hom\_50}, the performance of 
other algorithms is not satisfactory, urgently requiring more advanced algorithms.

\textbf{Sparse Team Reward Tasks}.
In these tasks, value-based algorithms generally outperform actor-critic-based algorithms. In the \textit{monster\_crisis\_easy} task, only QPLEX trains relatively quickly and stably. Other value-based algorithms require a larger number of episodes to exceed a win rate of 0.8 and do not perform well in the first 100,000 episodes. In \textit{monster\_crisis\_hard}, some algorithms do not perform well, but as a actor-critic-based algorithm, HATRPO performs better than expected. The limitation of 
the performance of HATRPO in this task may lie in its inability to explore the entire space, thus failing to ensure monotonic improvement. Therefore, in \textit{monster\_crisis\_hard}, there is an urgent need for more advanced algorithms.

\textbf{Multi-Team Gaming Tasks}. 
In the tasks where the engaging teams are driven by scripts, apart from the poor performance of MAPPO in all tasks, actor-critic-based algorithms are superior to value-based algorithms. Within actor-critic-based algorithms, HATRPO, as the algorithm with the most precise monotonic improvement, performs the best. It can stably learn the superior policies in both \textit{flag\_capture\_script} and \textit{flag\_capture\_2scripts}. This indicates that in these tasks, computing only the first-order approximation or using clip clipping like HAPPO is not the optimal solution. Among value-based algorithms, QPLEX and WQMIX are the two best performing algorithms. Among them, QPLEX trains slightly faster, indicating that in simple tasks with fewer agents, QPLEX is the fastest learning algorithm among its value-based counterparts.

In \textit{flag\_capture\_mappo}, actor-critic-based algorithms train relatively quickly and can achieve the high win rate within 50,000 episodes. On the contrary, value-based algorithms can achieve the high win rate only after 50,000 episodes. Apart from QMIX, which performed poorly, the other algorithms all performed well. In \textit{flag\_capture\_double\_mappo}, all actor-critic-based algorithms perform well. In value-based algorithms, only QPLEX can achieve the high win rate after a large number of episodes.


\begin{table}[h]
\centering
\caption{The result of engaging with teams driven by MAPPO. The data represents the average win rate within the corresponding range of episodes.}
\label{tab:multi-team result}
{\small 
\begin{tabular}{c|ccc|ccc}
\toprule
\multirow{2}{*}{\bf ALGORITHM}                    &\multicolumn{3}{c|}{\bf \textit{flag\_capture\_mappo}} &\multicolumn{3}{c}{\bf \textit{flag\_capture\_double\_mappo}} \\ 
                            &\!$0k\!\sim\!50k$\!&\!$50k\!\sim\!100k$\!&\! $100k\!\sim\!150k$\!&\!$0k\!\sim\!50k$\!&\! $50k\!\sim\!100k$\!&\!$100k\!\sim\!150k$\\ \midrule
MAPPO  
& $0.52\!\pm\!0.25$ & $0.56\!\pm\!0.09$ & $0.50\!\pm\!0.17$ & $0.71\!\pm\!0.12$ & $0.71\!\pm\!0.08$ & $0.78\!\pm\!0.17$ \\
HAPPO          
& $0.65\!\pm\!0.20$ & $0.67\!\pm\!0.17$ & $0.77\!\pm\!0.11$ & $0.67\!\pm\!0.08$ & $0.68\!\pm\!0.18$ & $0.71\!\pm\!0.17$ \\
HATRPO          
& $0.54\!\pm\!0.28$ & $0.67\!\pm\!0.16$ & $0.51\!\pm\!0.29$ & $0.61\!\pm\!0.26$ & $0.65\!\pm\!0.34$ & $0.77\!\pm\!0.12$ \\
QMIX              
& $0.11\!\pm\!0.07$ & $0.02\!\pm\!0.03$ & $0.01\!\pm\!0.02$ & $0.07\!\pm\!0.02$ & $0.65\!\pm\!0.34$ & $0.77\!\pm\!0.12$ \\
QTRAN           
& $0.71\!\pm\!0.06$ & $0.73\!\pm\!0.13$ & $0.78\!\pm\!0.06$ & $0.25\!\pm\!0.28$ & $0.17\!\pm\!0.24$ & $0.13\!\pm\!0.13$ \\
QPLEX             
& $0.66\!\pm\!0.21$ & $0.96\!\pm\!0.03$ & $0.97\!\pm\!0.02$ & $0.29\!\pm\!0.20$ & $0.73\!\pm\!0.09$ & $0.87\!\pm\!0.14$ \\
WQMIX  
& $0.39\!\pm\!0.18$ & $0.79\!\pm\!0.08$ & $0.76\!\pm\!0.19$ & $0.27\!\pm\!0.13$ & $0.07\!\pm\!0.05$ & $0.10\!\pm\!0.10$ \\
\bottomrule
\end{tabular}
} 
\end{table}










\begin{table}[ht]
\centering
\caption{Common hyperparameters used for MAPPO, HAPPO, and HATRPO in Unreal-MAP.}
\label{Tab: hyperparameters1}
{\small 
\renewcommand{\arraystretch}{1.0} 
\begin{tabular}{ccccccc}
\toprule
                            & MAPPO & HAPPO & HATRPO\\ \midrule
share parameter & True & True & True\\ 
hidden sizes  
& 128 & 128 & 128 \\
use feature normalization        
& True & True & True \\
use naive recurrent policy          
& False & False & False \\
actor learning rate         
& 0.001 & 0.001 & 0.001 \\
critic learning rate             
& 0.0005 & 0.0005 & 0.0005 \\
eps of optimizer           
& 0.00001 & 0.00001 & 0.00001 \\
weight decay            
& 0 & 0 & 0 \\
clip parameter   
& 0.2 & 0.2 & 0.2 \\
entropy coefficient  & 0.01 & 0.01 & 0.01 \\
coefficient for value loss     & 1 & 1 & 1 \\
gamma
& 0.99 & 0.99 & 0.99 \\
GAE lambda         & 0.95 & 0.95 & 0.95\\
use a fixed optimisation order & \textbf{--} & False & False\\ 
kl threshold & \textbf{--} & \textbf{--} & 0.01\\ 
\bottomrule
\end{tabular}
} 
\end{table}

\begin{table}[ht]
\centering
\caption{Common hyperparameters used for QMIX, QTRAN, QPLE, and WQMIX in Unreal-MAP.}
\label{Tab: hyperparameters2}
{\small 
\renewcommand{\arraystretch}{1.0} 
\begin{tabular}{cccccccc}
\toprule
                            & QMIX & QTRAN & QPLEX & WQMIX\\ \midrule
optimizer & adam & adam & adam & adam\\ 
learning rate             
& 0.001 & 0.001 & 0.001 & 0.001 \\
state compat        
& mean observation & mean observation & mean observation & mean observation \\
hidden sizes        
& 128 & 128 & 128 & 128 \\
hypernet-dimension          
& 64 & 64 & 64 & 64 \\
TD lambda             
& 0.6 & 0.6 & 0.6 & 0.6 \\
\bottomrule
\end{tabular}
} 
\end{table}

\begin{table}[ht]
\centering
\caption{Common hyperparameters used for the 15 tasks.}
\label{Tab: hyperparameters3}
{\small 
\renewcommand{\arraystretch}{1.0} 
\begin{tabular}{cccccccc}
\toprule
                            & Metal Clash & Flag Capture & Monster Crisis & Navigation Game\\ \midrule
simulation time step & 1/2560 s & 1/2560 s & 1/2560 s & 1/2560 s\\ 
simulation time interval             
& 1/2 s & 1/2 s & 1/2 s & 1/2 s \\
time dilation factor       
& 64 & 64 & 64 & 64 \\
parallel environment         
& 32 & 32 & 64 & 32 \\
maximum episode step       
& 125 & 250 & 100 & 150 \\
\bottomrule
\end{tabular}
} 
\end{table}





\end{document}